\documentclass[10pt,journal,cspaper,compsoc]{IEEEtran}
\pdfoutput=1
\usepackage{amsfonts}
\ifCLASSOPTIONcompsoc
\usepackage[nocompress]{cite}
\else
\usepackage{cite}
\fi
\usepackage{amsmath}
\usepackage{algorithm}
\usepackage{algorithmic}
\usepackage{stfloats}
\usepackage{url}
\usepackage{mathrsfs}
\usepackage{dsfont}
\usepackage{multirow}
\usepackage{multicol}
\usepackage{color}
\usepackage{graphics}
\usepackage{graphicx}
\usepackage{subfig}

\begin{document}
	
	\title{CTNet: Context-based Tandem Network for Semantic Segmentation}

	\author{Zechao Li, Yanpeng Sun, and Jinhui Tang
		\IEEEcompsocitemizethanks{\IEEEcompsocthanksitem Z. Li, Y. Sun and J. Tang are with School of Computer Science and Engineering, Nanjing University of Science and Technology, No 200 Xiaolingwei Road, Nanjing 210094, China. E-mail: \{zechao.li, yanpeng\_sun, jinhuitang\}@njust.edu.cn (Corresponding author: Jinhui Tang)}
	}
	\markboth{IEEE Transactions on PATTERN ANALYSIS AND MACHINE INTELLIGENCE,~Vol.~x, No.~x, Month~Year}{Z. Li, Y. Sun and J. Tang: CTNet: Context-based Tandem Network for Semantic Segmentation}
	
\IEEEcompsoctitleabstractindextext{
\begin{abstract}
Contextual information has been shown to be powerful for semantic segmentation. This work proposes a novel Context-based Tandem Network (CTNet) by interactively exploring the spatial contextual information and the channel contextual information, which can discover the semantic context for semantic segmentation. Specifically, the Spatial Contextual Module (SCM) is leveraged to uncover the spatial contextual dependency between pixels by exploring the correlation between pixels and categories. Meanwhile, the Channel Contextual Module (CCM) is introduced to learn the semantic features including the semantic feature maps and class-specific features by modeling the long-term semantic dependence between channels. The learned semantic features are utilized as the prior knowledge to guide the learning of SCM, which can make SCM obtain more accurate long-range spatial dependency. Finally, to further improve the performance of the learned representations for semantic segmentation, the results of the two context modules are adaptively integrated to achieve better results. Extensive experiments are conducted on three widely-used datasets, i.e., PASCAL-Context, ADE20K and PASCAL VOC2012. The results demonstrate the superior performance of the proposed CTNet by comparison with several state-of-the-art methods.
\end{abstract}
		
}

\maketitle
\IEEEdisplaynotcompsoctitleabstractindextext
\IEEEpeerreviewmaketitle

\section{Introduction}
\begin{figure}
	\centering
	\includegraphics[width=\linewidth]{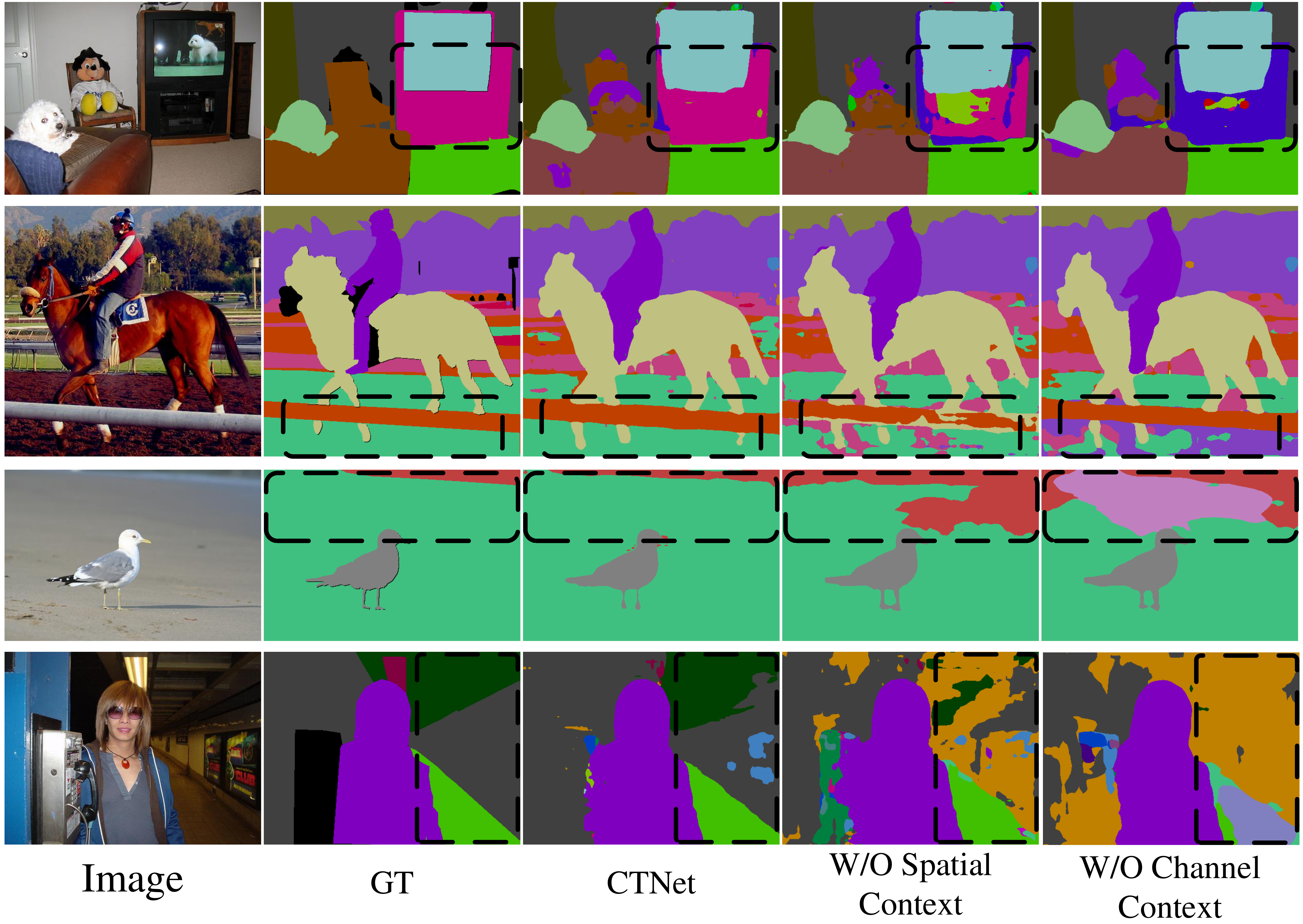}
	\caption{Semantic segmentation is to identify the category of each pixel. It is very challenging to parse pixels with similar appearance. It is prone to some regional segmentation errors without considering spatial context, while ignoring channel context is prone to incorrect category information. Areas in the black box are easily confused.}
	\label{fig1}
	\vspace{-4mm}
\end{figure}

\IEEEPARstart{S}{emantic} segmentation is a crucial but challenging task in the fields of multimedia and computer vision, which has been applied in various applications. The goal of semantic segmentation is to predict the semantic label for each pixel of the image. The main challenge arises from the difficulty to distinguish some confusing categories accurately with similar appearance. For example, as shown in Figure \ref{fig1}, the object of \textit{'chair'} has the same color to the object of \textit{'sofa'}, and the regions of \textit{'water'} and \textit{'ground'} are visually similar in color and shape. It is difficult to distinguish these regions by only considering the appearance. Therefore, contextual information has been widely explored for semantic segmentation to improve the segmentation performance \cite{hung2017scene, chen2017rethinking, ding2018context, zhao2017pyramid}.
	
\begin{figure*}
	\centering
	\includegraphics[width=0.97\textwidth]{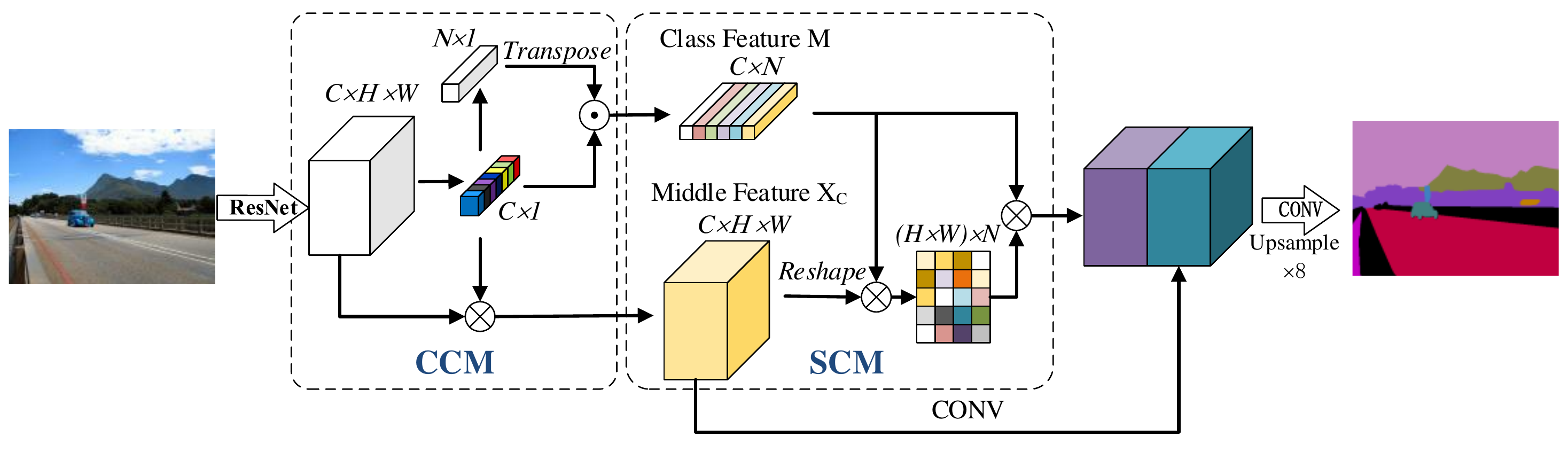}
	\caption{The illustration of the proposed CTNet framework. It jointly explores the spatial dependency and the semantic dependency by levering the Spatial Contextual Module (SCM) and the Channel Contextual Module (CCM). With the extracted feature maps by the pre-trained backbone, CCM explores the semantic dependencies to learn the new feature map and the feature representation of each category. The learned features by CCM are utilized by SCM to update the feature map by considering the spatial context.}
	\label{fig2}
	\vspace{-4mm}
\end{figure*}
	
Many methods have been proposed to explore the spatial contextual information based on the Fully Convolutional Network (FCN) framework \cite{shen2020ranet, xiao2018unified, zhang2020feature, seyedhosseini2015semantic}. The spatial context describes the relationship among pixels since each object in one image is described by many pixels. To explore the spatial dependence, some works \cite{ding2020semantic} have been studied to expand the network receptive field. Deeplabv3 \cite{chen2017rethinking} and PSPNet \cite{zhao2017pyramid} both use multi-scale feature extraction schemes to expand the spatial receptive field. However, these schemes only focus on local feature relationships, and the resulting contextual information is limited \cite{chen2014semantic}. Recently, the self-attention scheme brings in new ideas to capture long-term dependencies \cite{wang2018non}. Unfortunately, it leads to large computational and memory cost, which makes it unsuitable for semantic segmentation. Hence, some works improve the self-attention scheme for semantic segmentation. CCNet \cite{huang2019ccnet} and EMANet \cite{li2019expectation} adopt a sparse attention mechanism to reduce the computational complexity of the model without decreasing the network performance. These methods prove that pixels with the same category contribute the most to the spatial context. However, they do not take into account relationships between pixels and categories to directly construct the spatial context information. These relationships are helpful to not only reduce the noise information in the context, but also make the spatial context more interpretable. Meanwhile, these spatial context-based methods are easy to introduce category information that does not exist in the image during the segmentation process. As shown in the third image in Figure \ref{fig1}, there are no region in the image that belong to \textit{'sky'}, but the pixels belonging to \textit{'ground'} are recognized as \textit{'sky'} in the segmentation process. The occurrence of this phenomenon indicates that spatial contexts do not contain category information of images. Therefore, spatial contexts-based methods are easy to lose the field in the category dimension of images, this problem can be called category field deficiency.

Actually, the feature map of each channel corresponds to a specific semantic response \cite{fu2020contextual, ni2019raunet}. Different semantic responses are correlated to each other. Thus, the channel context should be explored to enhance useful features and suppress features that are less useful to the current task. Some works are proposed to obtain the channel context in feature map \cite{woo2018cbam, hu2018squeeze, li2019selective}. SENet adopts the Squeeze-and-Excitation operation to obtain the channel context \cite{hu2018squeeze}. SKNet takes advantage of the attention mechanism to get richer channel context \cite{li2019selective}. EncNet obtains global features through encoder layers and channel context through fully connected layers \cite{zhang2018context}. These methods treat each channel equally by only considering the interaction information between all channels. ECANet \cite{wang2020eca} is proposed to capture local cross-channel interactions. But the used single-scale ECA block has limited context information and is not suitable for downstream tasks. Meanwhile, These categories are easily confused in image segmentation based on channel context only. As shown in the third image of Figure \ref{fig1}, \textit{'ground'} and \textit{'water'} belong to the category of image, but  areas belonging to the \textit{'ground'} are wrongly classified as \textit{'water'}. Therefore, channel contexts-based methods are easy to lose the field in the pixel dimension of images, this problem can be called pixel field deficiency. 
	
Current work shows that channel context and spatial context are very important for segmentation, therefore how to use them reasonably is an urgent problem to be solved. Some methods are proposed to explore the channel context and spatial context \cite{fu2019dual, woo2018cbam}. DANet uses a parallel network to capture context on spatial and channel separately in two branches \cite{fu2019dual}. However, these methods deem that the channel context and the spatial context are independent, and do not explore the complementarity between them.
	
Towards this end, this work proposes a new Context-based Tandem Network (CTNet) leveraging the Channel Contextual Module (CCM) and the Spatial Contextual Module (SCM), which is illustrated in Figure \ref{fig2}, to interactively explore the channel and spatial contextual information. The proposed CTNet utilizes the rich semantic information encoded in the high-level feature maps of channels to guide the learning of the spatial dependency. It can guarantee that the proposed CTNet captures the spatial contextual dependency between pixels and the semantic dependency between channels simultaneously.

For CCM, Multi-local Channel Excitation (MCE) block is designed to explore the channel context by learning the semantic dependencies between multi-local channel feature maps, since the contextual information of local channels from different scales is complementary. Besides, a new class probability loss is developed to standardize the training, which enforces the network to better capture the channel context and accurately predict the probability of the class in the image. With the uncovered channel context, CCM generates a semantic representation vector (termed as the class feature) for each category by predicting its probability appearing in the image and a new feature map (termed as the middle feature) by updating each channel feature map. For SCM, a new self-attention mechanism is proposed to capture the spatial context. The global dependencies among all pixels are learned based on the middle feature and the class feature, and features are updated by aggregating the category feature based on the correlations between pixels and categories. Actually, the features learned by CCM can be deemed as the prior knowledge to guide the SCM learning, which can improve the features learned by SCM. Thus, CCM and SCM are trained interactively rather than independently. Finally, the learned feature maps by CCM and SCM are adaptively fused for semantic segmentation. The proposed CTNet can discover the complete long-range dependencies by interactively considering the spatial and channel relationships. The superior performance of CTNet for semantic segmentation is demonstrated in comparison with the state-of-the-art methods. Especially, 55.5\% mIoU on PASCAL-Context, 45.94\% mIoU on ADE20K and 85.3\% mIoU on PASCAL VOC2012 without any pre-training are achieved.
	
The main contributions of this paper are summarized as follows:
\begin{itemize}
\item We propose a novel Context-based Tandem Network (CTNet) by interactively exploring the spatial contextual dependency between pixels and the semantic dependency between channels. The joint learning of the long-range semantic and spatial dependencies can significantly improve the desired features for semantic segmentation. To our best knowledge, it is the first work to interactively leverage the spatial context and the channel context for semantic segmentation.
\item The Channel Contextual Module (CCM) leverages the proposed MCE block and the developed class probability loss to explore multi-scale local channel contexts and ensure that channel contexts can contain all categories and reflect the differences between different categories.
\item A new self-attention scheme in SCM is proposed to model the global spatial dependency by exploring the correlations between pixels and categories, which can reduce the computational complexity while guaranteeing the performance.
\end{itemize}
	
\section{Relate Work}
	
\subsection{Context for Segmentation}
Context information plays an important role for semantic segmentation. Many methods have been proposed to use the rich context to improve the segmentation performance. The previous methods could be roughly divided into two contextual dimensions: spatial dimension and channel dimension.
	
Spatial context methods \cite{chen2017rethinking, lin2017exploring, liu2015parsenet, zhang2020causal} aim to expend the receptive field of the network by using the spatial contextual information. Deeplabv2 \cite{chen2014semantic} and Deeplabv3 \cite{chen2017rethinking} both use the dilated convolution to expand the receptive field of the network and embed context information through atrous spatial pyramid pooling. PSPnet \cite{zhao2017pyramid} uses a multi-scale pyramid pooling module to collect context information at different scales. For the channel dimension, some works use adaptive strategies to obtain long dependencies between channels. EncNet \cite{zhang2018context} and SENet \cite{hu2018squeeze} use the fully connected layer's autonomous learning ability to acquire the long dependency of channel.
	
In order to achieve higher segmentation performance, some methods use these two contexts simultaneously. DANet \cite{fu2019dual} proposes a parallel framework to capture features dependencies in the spatial and channel dimensions respectively. CFNet \cite{zhang2019co} independently explores the spatial and channel contexts with two branches. One branch captures the relationship between pixel points using self-attention and the other branch captures channel context by convolution. The aforementioned methods explore the spatial context and the channel context individually or independently. There is no knowledge communicated between the spatial context and the channel context, which makes the performance limited. To address this problem, we propose to interactively exploring the spatial contextual dependency between pixels as well as the semantic dependency between channels, and exchange the uncovered knowledge.
	
\subsection{Self-attention Mode}
Attention mechanism \cite{huang2019interlaced, wang2018non, vaswani2017attention} is widely used in multimedia and computer vision since it can easily model long dependencies. Self-attention captures the context information of a location by weighted summation of all location information, to obtain the global perception field.
	
The self-attention scheme is first used in computer vision by the non-local methods and achieves good results \cite{wang2018non}. The current methods \cite{wang2018non, fu2019dual} based on the self-attention module mainly explore the non-local operations for image and video analysis. Unfortunately, the high computation and memory cost limits its application. Some improved methods have been proposed to address this problem \cite{huang2019ccnet, li2019expectation, chen20182, zhu2019asymmetric, yin2020disentangled}. $A^{2}$-Net \cite{chen20182} first selects some key factors from images and then indirectly obtains the long dependence between pixels by using the relationship between pixels and key factors. EMNet \cite{li2019expectation} uses the expectation-maximization iteration to obtain the correlation between pixels and key factors. These methods randomly select key factors such as pixels from images. These key factors are short of semantic information, which may be not good for semantic segmentation. Therefore, the proposed CTNet explores the category information of images to identify the key factors to promote the performance of semantic segmentation.
	
\section{The Proposed Approach}
This section will elaborate the proposed Context-based Tandem Network (CTNet) for semantic segmentation by jointly exploring the long-range semantic and spatial dependencies.
	
\subsection{Overview}
To jointly exploring the channel context and the spatial context, a novel tandem network including the Channel Context Module (CCM) and the Spatial Context Module (SCM) is proposed for semantic image segmentation, as illustrated in Figure \ref{fig2}. These two dimensions of contexts communicate with each other to improve the feature maps learned by each module. The proposed network finally aggregates the features of the two modules to obtain more accurate feature representations for semantic segmentation.
	
The proposed CTNet explores the channel context and the spatial context in tandem. The pre-trained deep convolution network is utilized as the backbone to extract the initiate feature maps, which is used as the input of CCM. To keep more details in the feature map, the size of the initiate feature maps is set to 1/8 of the input image size. CCM is proposed to learn the channel context vector, which can characterize the importance of each channel. Then, the initiate feature map is updated by multiplying the channel context vector to obtain a new feature map, termed as the middle feature in this paper. Besides, the channel context vector contains rich semantic information since the feature map of each channel corresponds to a specific semantic response. Thus, the probability of each category appearing in the image is predicted by using the channel context map. CCM learns the class-specific feature representation for each category (termed as the class feature) by multiplying this probability with the channel context vector. The learned middle feature and class feature are used as the input to SCM. They can be deemed as the prior knowledge to guide the SCM learning. To explore the long-range spatial context, SCM introduces a new self-attention model. The spatial relationships between pixels and categories are modeled, and then aggregated by using the class features to obtain the characteristic representation of the long-dependence relationship. The communication between CCM and SCM can make them compatible each other. Finally, feature maps of these two modules are integrated to generate better feature representations. The segmentations results are obtained by introducing the final convolution layer and a series of up-sampling operations.
	
\begin{figure}
\centering
\includegraphics[width=\linewidth]{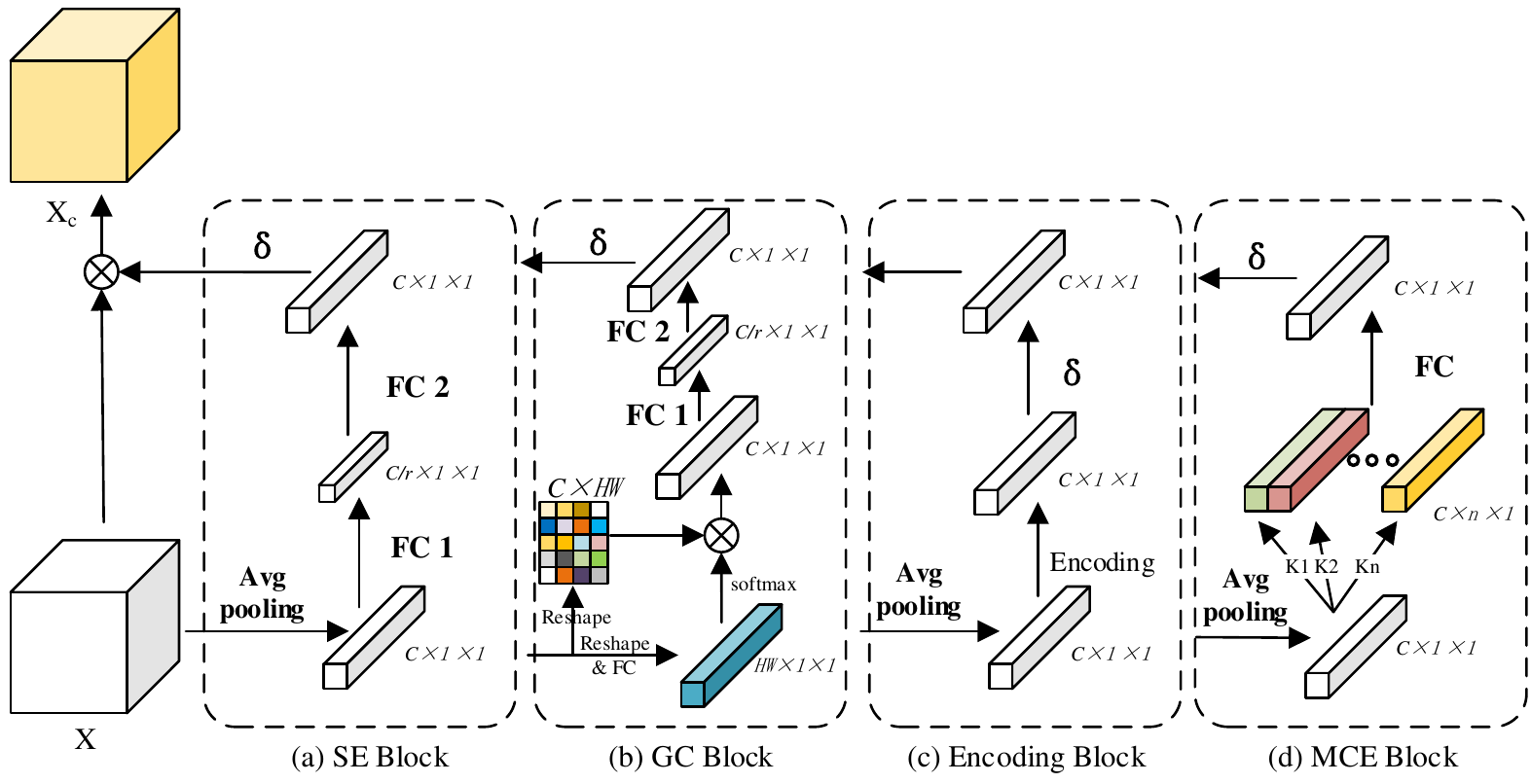}
\caption{Comparison among different channel context extraction models. (a) Squeeze Excitation (SE) block; (b) Global Context (GC) block; (c) Encoding block; (d) The proposed Multi-local Channel Excitation (MCE) block. The shape of the feature map $x$ and $X_{c}$ is $C\times H\times W$, $\otimes$ represents the channel-wise multiplication.}
\label{fig3}
\vspace {-4mm}
\end{figure}
	
\subsection{CCM Module}
Each channel feature map corresponds to a specific semantic response in the high-level feature, and different semantic responses are related to each other. Using channel context, that is, the dependencies between channels, the feature representation of specific semantics can be improved and the feature mapping is re-calibrated. Existing methods extract channel context by designing different blocks shown in Figure \ref{fig3}. SE block \cite{hu2018squeeze}, GC block \cite{cao2019gcnet} and Encoding block \cite{zhang2018context} improve the representational ability of models by capturing global channel interaction information. However, since each channel is only closely related to its neighboring channels, these interaction information are redundant \cite{wang2020eca}.
\begin{figure}[t]
\centering
\includegraphics[width=\linewidth]{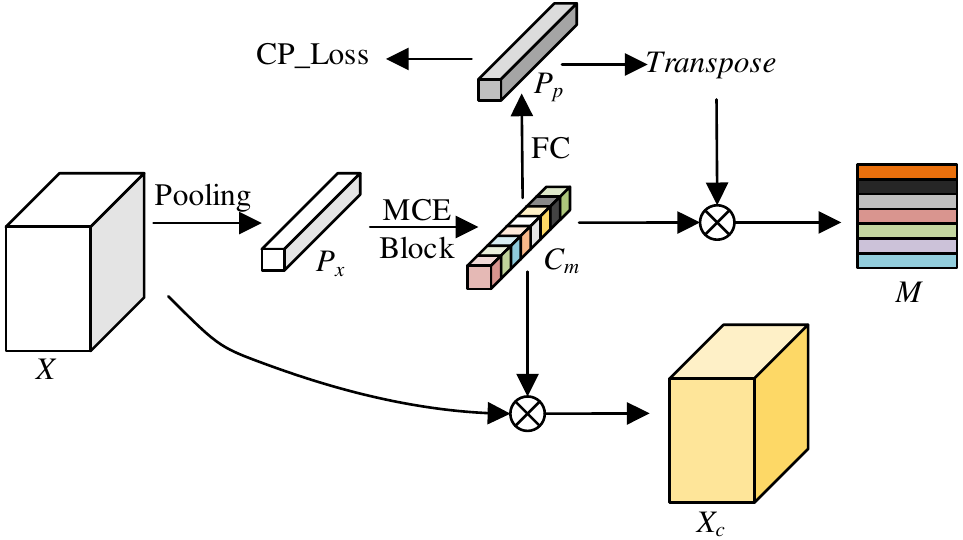}
\caption{The details of the Channel Context extraction Model (CCM).}
\label{fig4}
\vspace{-4mm}
\end{figure}
	
Toward this end, the Multi-local Channel Excitation (MCE) block is proposed by fusing multi-scale local channel contexts, where different scales represent different numbers of neighbor channels associated with the current channel. The structure of the MCE block is shown in Figure \ref{fig3} (d). Given a feature map $X\in R^{C\times H\times W}$ as input, the global pooling operation is employed to turn each two-dimensional feature channel into a real number, which has a global receptive field to some extent. That is, a rough global information $P_{x}\in R^{C\times 1\times 1}$ that  represents the global distribution of responses on the characteristic channel is obtained. Then the MCE block introduces the 1-D convolution to capture the local channel context from the rough global information.
	
To capture more comprehensive channel context information, the MCE block first explores the local channel context of different scales by adopting multiple 1-D convolutions with different kernels, and then aggregates them to obtain the multi-scale local channel context map $C_{c}\in R^{C\times s\times 1}$.
\begin{align}
{C_c}&= Concat\left ( C_{1},C_{2},...,C_{s} \right )\notag\\
&=Concat\left ( f_{k_1}\left ( P_{x} \right ),f_{k_2}\left ( P_{x} \right ),...,f_{k_s}\left ( P_{x} \right ) \right)
\end{align}
where $s$ is the number of the 1-D convolutions, $Concat$ denotes the Concatenation operation, and $f_{k_{i}}$ is the 1-D convolution with the kernel $k_{i} (i = 1,...,s)$. Then, the MEC block adopts the full connection layer to extract the final channel context $C_{m}\in{R}^{C\times 1\times 1}$ from the aggregated channel map $C_{c}$.
\begin{equation}
	\\\begin{array}{l}
	C_{m}= \sigma \left ( W_{1}C_{c}+ b_{1} \right )
	\end{array}
\end{equation}
where $\sigma$ is the sigmoid activation function. $W_{1}\in R^{s\times 1}$ and $b_{1}\in R^{C\times 1}$ denote the parameter matrix and the bias vector of the full connection layer, respectively.
	
CCM is developed based on the MCE block to update the representation of the feature map with the channel context and get the class feature of the image, which is shown in Figure \ref{fig4}. That is, a new feature map called the middle feature $X_{c}\in R^{C\times H\times W}$ is generated by channel-wise multiplication between $C_{m}$ and the original feature map $X$.
\begin{align}
	{X_c} = {C_m} \otimes X
\end{align}
where $\otimes$ denotes the channel-wise multiplication.
	
High-level features always lose the information of small objects in the image, which makes it difficult for semantic segmentation tasks to recognize small objects. Meanwhile, to get a class feature map, CCM should learn the differences between categories in each image. For this purpose, a new loss function (Class Probability Loss, CP-Loss) is developed to regularize the training by making CCM predict the occurrence probability of object categories in the image. The CP-Loss is defined as follows.
\begin{align}
	l_{cp}(P_{p},P_{gt})=-\omega _{n}[P_{gt}\log P_{p}+(1-P_{gt})\log (1-P_{p})]
\end{align}
where $P_{p}$ is the predicted probability of categories in images, $P_{gt}$ is the target probability, and $\omega _{n}$ is the weight of the current batch. $P_{p}$ is learned through a fully connected layer with a parameter of $C\times N$ by $C_{m}$, where $N$ is the number of all the categories in the dataset. Actually, there is no information about the target probability of object categories appearing in the image. Thus, $P_{gt}$ is calculated based on the ground-truth annotation. The target probability of category $i$ in the image is obtained by $P_{gt}^{i}=\frac{P^{i}}{\sum_{i=1}^{N}P^{i}}$, where $P^{i}$ is the occurrence frequency of pixels belonging to category $i$ in the current image. CP-Loss ensures that the channel context can contain not only all the categories in the image, but also the occurrence probability of all the categories in the image.
	
Based on the channel context $C_{m}$ and the probability $P_{p}$ of each category, a class feature matrix $M$ composed of all the objects feature vectors in the image can be obtained by
\begin{align}
	M = {C_m} \odot {{P_p}^{T}}
\end{align}
where $\odot$ represents the cross product. The middle feature map $X_{c}$ and the class feature map $M$ are input into the SCM module. The feature map provided by CCM can be regarded as a prior knowledge to guide the SCM learning, which can improve the characteristics of SCM learning. At the same time, the class feature matrix provides an independent feature representation for each category in the image, which can help SCM capture the spatial context more concisely.
	
\begin{figure}
\centering
\includegraphics[width=\linewidth]{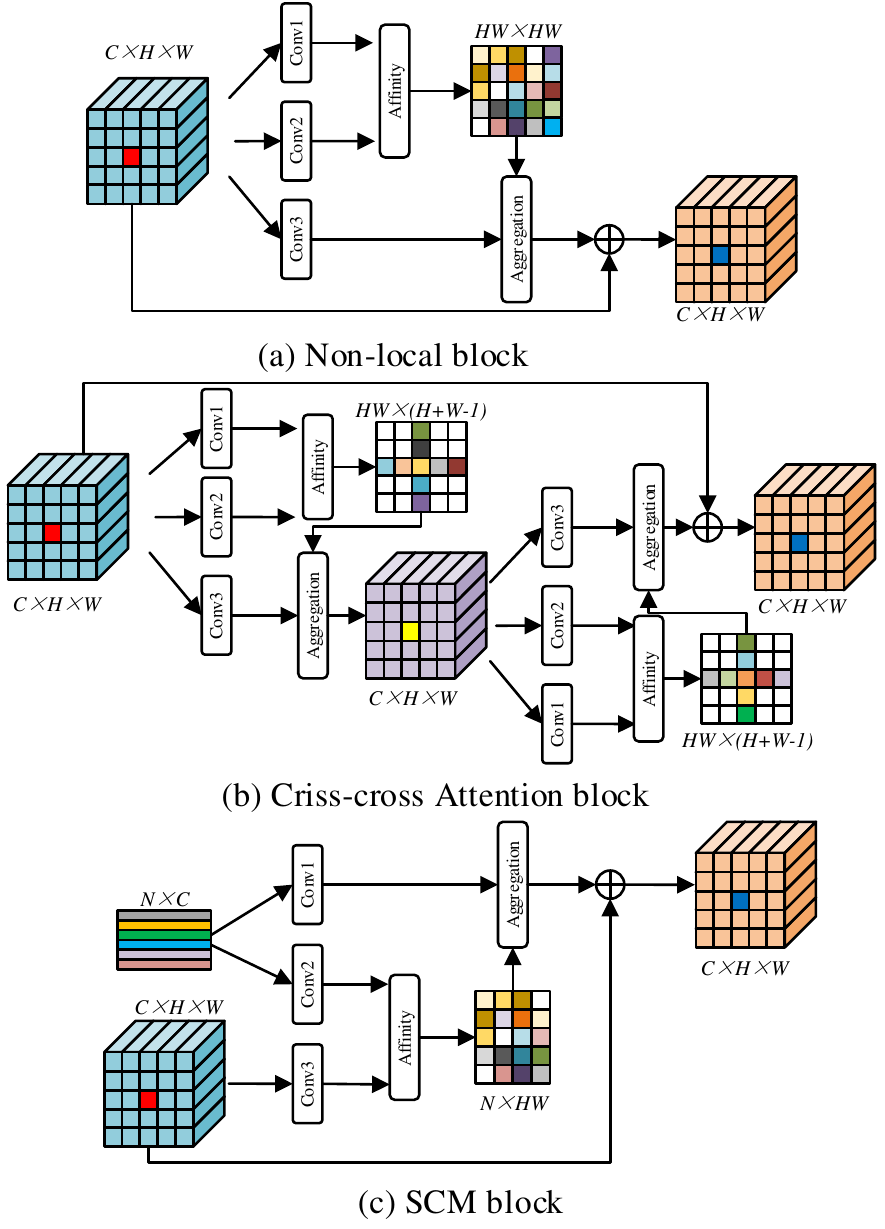}
\caption{Comparisons with different spatial context extraction models based self-attention. (a) Non-local block; (b) Criss-cross attention block; (c) The proposed SCM block.}
\label{fig5}
\vspace{-4mm}
\end{figure}
	
\subsection{SCM module}
To improve the segmentation performance, it is necessary to capture the long-term dependent context information in spatial dimension. The self-attention based methods are proposed to capture global spatial contexts by enlarging the receptive field of the model \cite{zhao2017pyramid, chen2017rethinking, lin2018multi}. As shown in Figure \ref{fig5} (a), the non-local block \cite{wang2018non} directly calculates correlations between pixels to capture the spatial context, which leads to high computational complexity and large memory consumption. The Criss-Cross attention block \cite{huang2019ccnet}, shown in Figure \ref{fig5} (b), is designed to improve non-local block by repeatedly calculating the correlation of pixels in the direction of Criss-Cross. However, it still describes the spatial context based on correlations between pixels. To this end, a new self-attention scheme is proposed in the SCM module as shown in Figure \ref{fig5} (c). The proposed SCM block explores the spatial context by calculating the relationship between the feature map and the higher-level semantic representation in images (e.g., the class feature).
\begin{figure}[t]
\centering
\includegraphics[width=\linewidth]{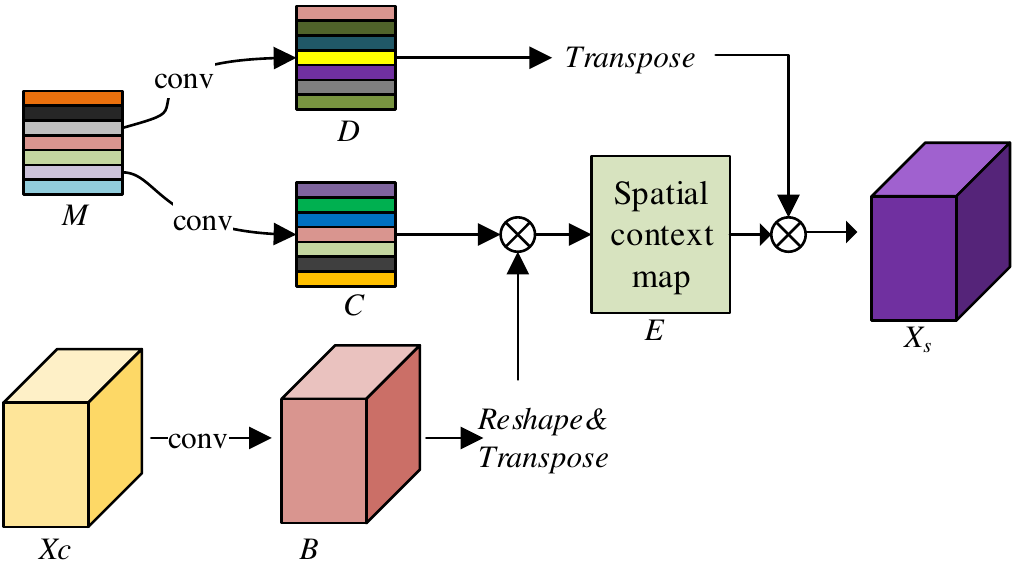}
\caption{The details of the Spatial Context extraction Model (SCM).}
\label{fig:scm}
\vspace{-4mm}
\end{figure}
	
To effectively the reduce computational complexity without affecting performance, SCM transforms the problem of learning the spatial context into the problem of matching the category of each pixel on the feature map, and then describes the global spatial context using the correlation between pixels and categories. The structure of the SCM module is shown in Figure \ref{fig:scm}. To make spatial context and channel context information interactive, the middle feature map $X_{c}$ and category feature representation matrix $M$ are input into the SCM module. First, two new feature representation $B\in R^{C/s\times H\times W}$ and $C\in R^{C/s\times N}$ are respectively obtained based on $X_{c}$ and $M$ by using the convolution layer to further reduce the computation cost, where $s$ is the reduction ratio. Then $B$ is reshaped to $R^{C/s\times HW}$. The spatial context map $E\in R^{HW \times N}$ is obtained by using a softmax layer.
\begin{equation}
	\\{e_{ij}} = \frac{{\exp \left( {B_i^{T} \cdot {C_j}} \right)}}{{\sum\nolimits_{k = 1}^N {\exp \left( {B_i^{T} \cdot {C_k}} \right)} }}
\end{equation}
where $e_{ij}$ is the correlation score between the $i$-th pixel and the $j$-th semantic category. It not only acquires the matching relationship between pixels and categories, but also aggregates pixels with the same matching relationship to indirectly obtain correlations between pixels.
	
To further improve the feature representation of pixels, we re-calibrate the pixel feature representation using the class feature representation matrix $M$. Specifically, $M$ is input into a new $1\times 1$ convolution layer to generate a new feature map $D\in R^{C/s\times N}$. Then, we obtain the feature map $X_{s}\in R^{C\times H\times W}$ updated with the spatial context through a $1\times 1$ convolution layer.
\begin{align}
X_S = \rho \left( {\sum\nolimits_{k = 1}^N {\left( {E_k^{T} \cdot {D}} \right)} } \right)
\end{align}
where $\rho$ denotes the transformation function composed of $1\times 1 \,conv\rightarrow BN\rightarrow ReLU$. It can be inferred that $X_S$ at each position is a weighted sum of the features across spatial context matrix and the category feature matrix. Therefore, similar semantic features achieve mutual gains through the category feature matrix. When the feature map is $C\times H\times W$, the computational complexity of the SCM block is $\mathcal{O} \left ( HW\times N \times C\right )$ while the computational complexity of the non-local block is $\mathcal{O} \left ( HW\times HW \times C\right )$, where $N << HW$. That is, the SCM block can not only address the problem of the non-local block, but also provide a new direction to describe the spatial context.
	
\subsection{Loss Function}
In order to take full advantage of the context in both dimensions, the feature maps from the two modules are integrated. The middle features generated by the CCM is fed into the convolution layer for feature extraction. And then the extracted feature is connected with the feature map generated by the SCM module. The $1\times 1$ convolution layer is introduced to integrate the connected feature map for generating the final feature map. The segmentation result is obtained by up-sampling the final feature map to the original image size with the bilinear interpolation method. To evaluate the segmentation performance, the segmentation loss $l$ is introduced by measuring the difference between the segmentation result and the ground truth.
	
In order to enhance the representation capability of the network, we introduce the auxiliary supervision to improve the performance of segmentation. Following the previous work \cite{chen2017rethinking, zhang2018context, zhu2019asymmetric}, the auxiliary loss $l_{au}$ is added after ResNet-4 stage by an additional FCN head. Besides, the developed CP-Loss $l_{cp}$ is introduced to the restrict network training.
	
To jointly explore the above terms, the final objective function of the proposed CTNet contains the segmentation loss $l$, the auxiliary loss $l_{au}$ and the CP-loss $l_{cp}$.
\begin{align}
L = l + \alpha  \cdot {l_{au}} +  \beta \cdot {l_{cp}}
\end{align}
where $\alpha$ and $\beta$ are the trade-off parameters corresponding to auxiliary loss and CP-loss, respectively. In other work \cite{huang2019ccnet, li2019expectation, yuan2020object}, the auxiliary loss is the only additional loss, and the weight of additional loss is set to $0.4$. Following the setting, we set $\alpha+\beta=0.4$, and default $\alpha=0.3$, $\beta=0.1$ in experiments.
	
\section{Experimental}
\label{exp}
To evaluate the proposed CTNet network, extensive experiments are conducted on three widely-used datasets, i.e., PASCAL-Context \cite{mottaghi2014role}, ADE20K \cite{zhou2017scene} and PASCAL VOC2012 \cite{everingham2010pascal}. In this section, we first introduce the used datasets and implementation details, then verify the superiority of the model by the ablation study on PASCAL-Context, and finally show the results of the network on these three datasets. All the results are obtained by the proposed CTNet without using COCO pre-training.
	
\subsection{Datasets and Evaluation Metrics}
Experiments are conducted on the PASCAL-Context \cite{mottaghi2014role}, ADE20K \cite{zhou2017scene} and PASCAL VOC2012 \cite{everingham2010pascal} datasets.
\begin{itemize}
\item \textbf{PASCAL-Context} \cite{mottaghi2014role} is a challenging semantic segmentation dataset composed of images from PASCAL VOC2010. It has $4,998$ images in the training set and $5,105$ images in the test set. There are totally $60$ categories including $59$ categories of objects and the background category.   
\item \textbf{ADE20K} \cite{zhou2017scene} is issued by MIT, which can be used for scene perception, analysis, segmentation, etc. It has $20,000$ images in the training set, $2,000$ images in the validation set and $3,000$ images in the test set, respectively. It contains $150$ semantic categories.
\item \textbf{PASCAL VOC2012} \cite{everingham2010pascal} is one of the standard benchmarks in the field of semantic segmentation. It contains $21$ categories including one background class. The original dataset has $1,464$ images in the training set, $1,449$ images in the verification set and $1,456$ images in the test set. The augmented annotation set \cite{hariharan2015hypercolumns}, which expanded the training set from the original data to $10,582$ images, is utilized.
\end{itemize}
	
\begin{figure}
\centering
\includegraphics[width=\linewidth]{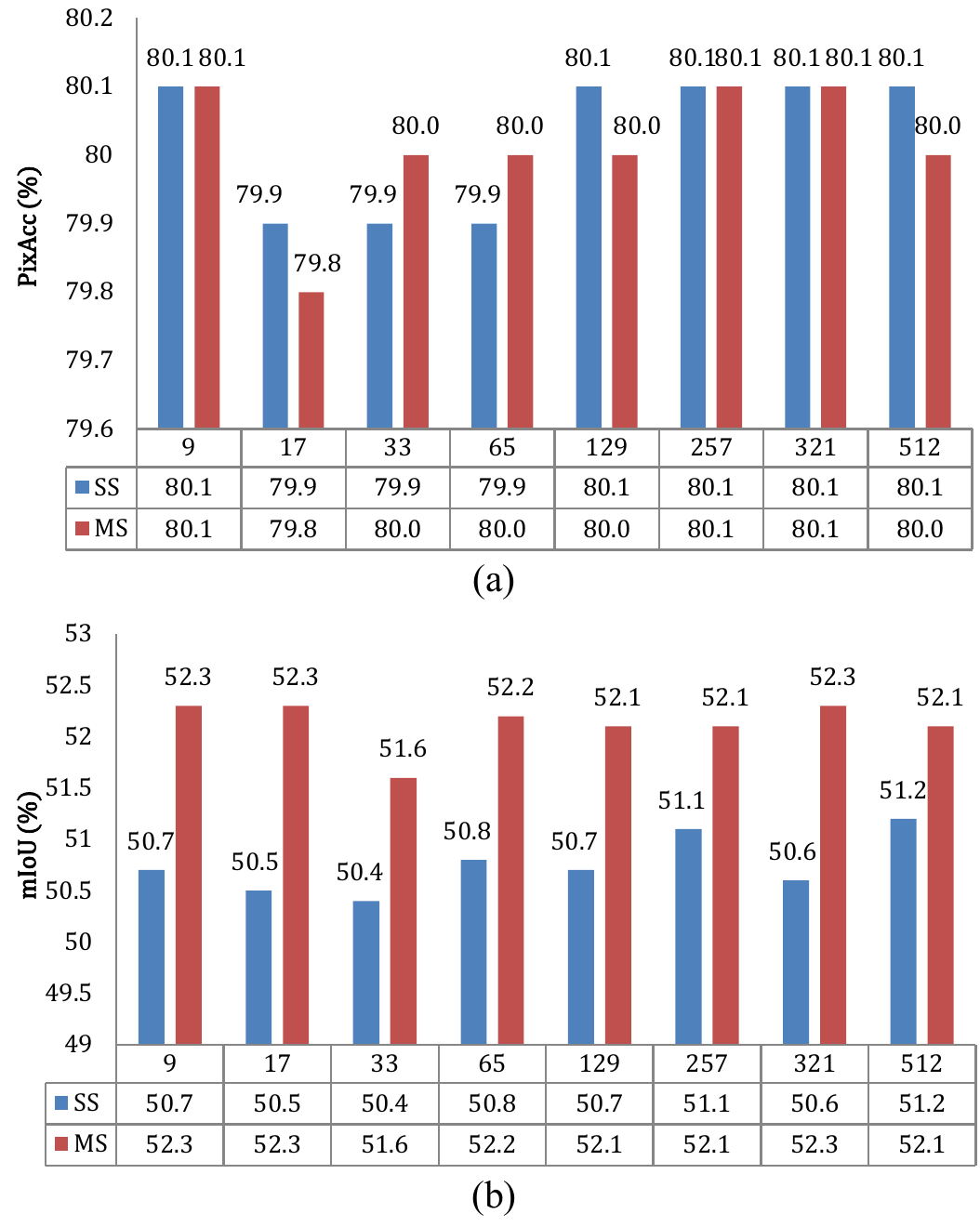}
\caption{Ablation study of different scale local channel contexts in CTNet on PASCAL-Context. (The horizontal coordinates represent the kernel size of 1-D convolution.)}
\label{fig7}
\end{figure}
	
Following previous work \cite{huang2019ccnet, fu2019dual, zhang2019co}, the multi-scale (MS) test is adopted by adjusting the original image to different scales ${0.5, 0.75, 1.0, 1.25, 1.5, 1.75}$. The scaled images are resized to the size of training images and fed into the network. The final result is obtained by averaging the network predictions in multiple scales. To evaluate the performance of semantic image segmentation methods, the widely-used metric mIoU (mean IoU, mean of class-wise intersection over union) is introduced. Besides, in ablation study, the mIoU and PixAcc (pixel accuracy) metrics are used.
	
\subsection{Implementation Details}
\subsubsection{Network}
The experimental system is based on Pytorch. The ImageNet pre-trained ResNet \cite{he2016deep} is utilized as the backbone of CTNet. CTNet uses ResNet-50 as the backbone in ablation study to speed up experiments, and ResNet-101 as the backbone in other experiments. In order to solve the problem of the memory consumption caused by dilated convolution and improve the efficiency of the segmentation, we use the JPU \cite{wu2019fastfcn} module rather than the dilated convolution strategy \cite{huang2019ccnet, fu2019dual, li2019expectation} to ensure that the feature map is $1/8$ of the input image. At the same time, we train CTNet with synchronized BN \cite{wu2019fastfcn}.
	
\subsubsection{Training}
Following the previous work \cite{chen2017rethinking, zhang2019co}, the SGD method is utilized to optimize the network. The momentum is set to $0.9$ and the weight decay is set to $0.0001$.  A poly learning rate policy is adopted, where the initial learning rate is multiplied by $(1-\frac{iter}{total-iter})^{power}$ with $power=0.9$. For the ADE20K dataset, the initial learning rate is set to $0.005$ with $300$ training epoch, while the initial learning rate is set to $0.001$ with $250$ training epoch for the other datasets. The batch size is set to $16$ for all the datasets by following previous works \cite{wu2019fastfcn, zhang2019co, yuan2020object}. For data augmentation, we only randomly flip and scale the image in the range of $[0.5, 2]$, then randomly rotate the image between $[-10, 10]$, and finally use zero padding to fix the image to a uniform size when necessary.
	
\begin{table}[t]
\centering
\caption{Ablation study of the proposed MCE block on the PASCAL-Context dataset.}
\label{table1}
\renewcommand\arraystretch{1.3}
\begin{tabular}{c|c|c||c}
			\hline
			Method & Backbone & kernel size set $K$ & mIoU ($\%$)  \\
			\hline
			CTNet+SE \cite{hu2018squeeze} & ResNet-50 & - & 51.8 \\
			
			CTNet+ECA \cite{wang2020eca} & ResNet-50 & [9] & 52.2 \\
			\hline
			CTNet+MCE & ResNet-50 & [9, 17] & 52.4 \\
			
			CTNet+MCE & ResNet-50 & [9, 17, 33] & 52.3 \\
			
			CTNet+MCE & ResNet-50 & [9, 17, 33, 65] & \textbf{53.3} \\
			
			CTNet+MCE & ResNet-50 & [9, 17, 33, 65, 129] & 52.4 \\
\hline
\end{tabular}
\end{table}

\begin{table}[t]
\centering
\caption{Ablation study of the global pooling strategy in CCM on PASCAL-Context. \textbf{\textit{Max}} denotes the global maximum pooling strategy, \textbf{\textit{Avg}} is the global average pooling strategy, while \textbf{\textit{Mix}} represents the addition of Max and Avg.}
\label{table2}
		\renewcommand\arraystretch{1.3}
		\centering
		\begin{tabular}{c|c||c||c}
			\hline
			Method &Backbone &PixAcc ($\%$) &mIoU ($\%$)\\
			\hline
			FCN \cite{shelhamer2017fully}  &   ResNet-50&   76.3 &46.3\\
			EncNet \cite{zhang2018context} &ResNet-50 & 79.4& 49.2\\
			CFNet \cite{zhang2019co}& ResNet-50& 79.3& 51.6\\
			\hline
			CTNet+\textit{Max} & ResNet-50& 80.2& 52.3\\
			CTNet+\textit{Mix} & ResNet-50& 80.2& 52.5\\
			CTNet+\textit{Avg} & ResNet-50& \textbf{80.4}& \textbf{53.3}\\
			\hline
	\end{tabular}
\end{table}
\subsection{Ablation Study}
We first conduct a series of ablation study on the Pascal-Context val dataset to verify the effectiveness of CCM and SCM module, and then indicate the correctness of the model connection scheme in CTNet. Finally, experiments are designed to verify the necessity and effectiveness of the proposed CP-Loss. It is worth noting that mIoU on $59$ object categories is reported on Pascal-Context during ablation study by following previous work.
	
\subsubsection{For CCM}
In the CCM module, the MCE block is proposed to integrate the multi-scale local channel contexts, which is important for semantic segmentation. Thus, the ablation study is conducted to show the effectiveness of the proposed MCE block. All experiments are conducted by using ResNet-50 as the backbone.

First, experiments are conducted to validate the effectiveness of the used local channel context. For this purpose, the size of the convolution kernel $K$ is tuned within $\{9, 17, 33, 65, 129, 257, 321, 512\}$ by replacing the MEC block in the CCM module with the 1-D convolution. It is worth noting that the number of channels in the feature map is $512$. That is, the case when $K=512$ is corresponding to the global channel context. Besides, the results by using the multi-scale test (MS) and single-scale test (SS, directly feeding the original image into the network) are presented. The results in terms of PixAcc and mIoU are presented in Figure \ref{fig7}. From the results, it can be observed that the local channel context, such as $K=9$, achieves the competitive performance compared with the global channel context. However, the computational cost of exploring the global channel context is much higher than the local channel context. Besides, the results in terms of mIoU by using the multi-scale test are superior to ones by using the single-scale test. Therefore, it is reasonable and necessary to explore the local channel context and utilize the multi-scale input.
	
\begin{table}[t]
\caption{Comparison of computational cost on the Pascal-Context dataset. The Flops (G), Params (MB) and Memory (MB) are calculated with the input size of $513 \times 513$. }
\label{table3}
		\renewcommand\arraystretch{1.3}
		\centering
		\begin{tabular}{c|c|c|c}
			\hline
			Method &$\Delta$Flops&$\Delta$Params &$\Delta$Memory\\
			\hline
			PSPNet \cite{zhao2017pyramid}  & 79.88&23.09 &42.65\\
			Deeplabv3 \cite{chen2017rethinking}& 65.95& 16.13&91.49\\
			Deeplabv3+ \cite{chen2018encoder} & 85.25& 16.89&1250.78\\
			\hline
			OCNet \cite{yu2020context} &66.60&15.23&281.27 \\
			DANet \cite{fu2019dual}&  101.23 &23.96& 195.98\\
			CFNet \cite{zhang2019co} & 42.78 & 13.14& 179.37\\
			CCNet \cite{huang2019ccnet}& 101.13 & 23.94 & 93.54\\
			\hline
			OSCM &\textbf{29.63}&\textbf{7.11}&\textbf{88.74}\\
			CTNet &\textbf{29.64} & \textbf{7.11}& \textbf{88.74}\\
			\hline
	\end{tabular}
\end{table}
	
\begin{table}[!t]
\caption{Compared results in terms of mIoU on Pascal-Context and ADE20K.}
\label{table4}
\renewcommand\arraystretch{1.3}
		\centering
		\begin{tabular}{c|c|c|c}
			\hline
			Method &Backbone&Pascal-Context &ADE20k\\
			\hline
			PSPNet \cite{zhao2017pyramid}  & ResNet-101&47.8 &43.51\\
			\hline
			DANet \cite{fu2019dual}&  ResNet-101 &50.5& -\\
			CFNet \cite{zhang2019co} & ResNet-101 & 54.0& 44.89\\
			EMANet \cite{li2019expectation}& ResNet-101& 53.1& -\\
			CCNet \cite{huang2019ccnet}& ResNet-101 & - & 45.22\\
			\hline
			OSCM &ResNet-101&54.6&-\\
			CTNet &ResNet-101 &\textbf{56.2}& \textbf{45.94}\\
			\hline
\end{tabular}
\end{table}
	
Besides, the MCE block is developed by fusing multi-scale local channel contexts to obtain more accurate channel context. Experiments are conducted to verify its validity by comparing it with the ECA block \cite{wang2020eca} and the SE block \cite{hu2018squeeze}. The experimental results are shown in Table \ref{table1}. For MCE, different combinations of the scale sets are utilized, i.e., $K=[9, 17]$, $K=[9, 17, 33]$, $K=[9, 17, 33, 65]$, and $K=[9, 17, 33, 65, 129]$. From the results, it can be observed that the MEC block and ECA block perform better than the SE block, which can indicate that the local channel context is better for semantic segmentation that the global channel context. Second, the MCE block achieves better performance than the ECA block and the SE block. Because fusing multi-scale local channel contexts can capture better context relationships than the single-scale context. The best performance $52.9\%$ mIoU is achieved when the scale set is $K=[9, 17, 33, 65]$. However, the performance slightly decreases when unceasingly fusing the new local context since too many fused local channel contexts lead to the redundant channel context information.
	
In order to verify that features extracted by CCM contain the important information in the feature map, the performance of the global pooling strategy is evaluated. There are three pooling strategies, i.e., maximum pooling, average pooling, and mix pooling. the mix pooling is the addition of features from the maximum pooling and average pooling. The performance in terms of PixAcc and mIoU is shown in Table \ref{table2}. It can be easily seen that the proposed CTNet performs better than the current optimal partitioned network regardless of any global pooling strategy. Besides, CTNet with the global average pooling strategy achieves the best results, i.e., $80.4\%$ accuracy and $53.3\%$ mIoU on the PASCAL-Context dataset with ResNet-50 as the backbone. Thus, the global average pooling strategy is utilized in experiments.
	
\begin{table}[t]
	\caption{Ablation study of CCM and SCM on PASCAL-Context.}
	\label{table5}
		\renewcommand\arraystretch{1.3}
		\centering
		\begin{tabular}{c|c|c|c|c|c}
			\hline
			Method & Backbone &CCM & SCM&SS &MS\\
			\hline
			OCCM&ResNet-50  & $\textbf{\checkmark}$& &51.3 &52.6\\
			OSCM&ResNet-50& & $\textbf{\checkmark}$& 50.9&52.3\\
			CTNet&ResNet-50& $\textbf{\checkmark}$&$\textbf{\checkmark}$&51.7&\textbf{53.3}\\
			\hline
			OCCM&ResNet-101  & $\textbf{\checkmark}$& &54.0 & 55.6\\
			OSCM&ResNet-101& & $\textbf{\checkmark}$& 53.2 & 54.5\\
			CTNet&ResNet-101& $\textbf{\checkmark}$&$\textbf{\checkmark}$&54.4&\textbf{56.2}\\
			\hline
	\end{tabular}
\end{table}
	
\subsubsection{For SCM}
To reduce the extremely high computational cost caused by self-attention, the SCM module based on a new self-attention scheme is proposed by utilizing the relationship between pixels and categories to describe the global spatial context.
	
First, the proposed method is compared with traditional multi-scale methods including PSPNet \cite{zhao2017pyramid}, Deeplabv3 \cite{chen2017rethinking} and Deeplabv3+ \cite{chen2018encoder}, as well as the self-attention-based methods including OCNet \cite{yu2020context}, DANet \cite{fu2019dual}, CFNet \cite{zhang2019co}, EMANet \cite{li2019expectation} and CCNet \cite{huang2019ccnet}. Besides, the simplified version of the proposed CTNet, termed as OSCM, which only uses the SCM module as the segmentation head by removing the CCM module from CTNet, is also compared. The compared results in terms of the increment of GFLOPs, GPU memory cost and the parameter number are presented in Table \ref{table3}. It can be easily observed that the designed SCM module is a more lightweight self-attention module by taking up a small amount of computing resources with only 7.11MB parameters on the Pascal-Context dataset. It should be noted that the computational cost token by OSCM and CTNet are the same since the computational resource consumed by the CCM module is very small and has little effect on CTNet resource consumption.
	
To demonstrate the effectiveness of the SCM module, the performance is evaluated on the Pascal-Context and ADE20K datasets. For fair comparison, ResNet-101 is used as the backbone network. Table \ref{table4} shows the compared results in terms of mIoU. We can see that OSCM achieves 54.6\% mIoU on the Pascal Context dataset, which surpasses other methods. It demonstrates the effectiveness of SCM for semantic image segmentation by jointly exploring the channel context and spatial context. Besides, the proposed CTNet achieves 56.1\% and 45.94\% mIoU on Pascal-Context and ADE20k, respectively, which can also show the importance of the channel context.
	
\begin{figure}
\centering
	\includegraphics[width=\linewidth]{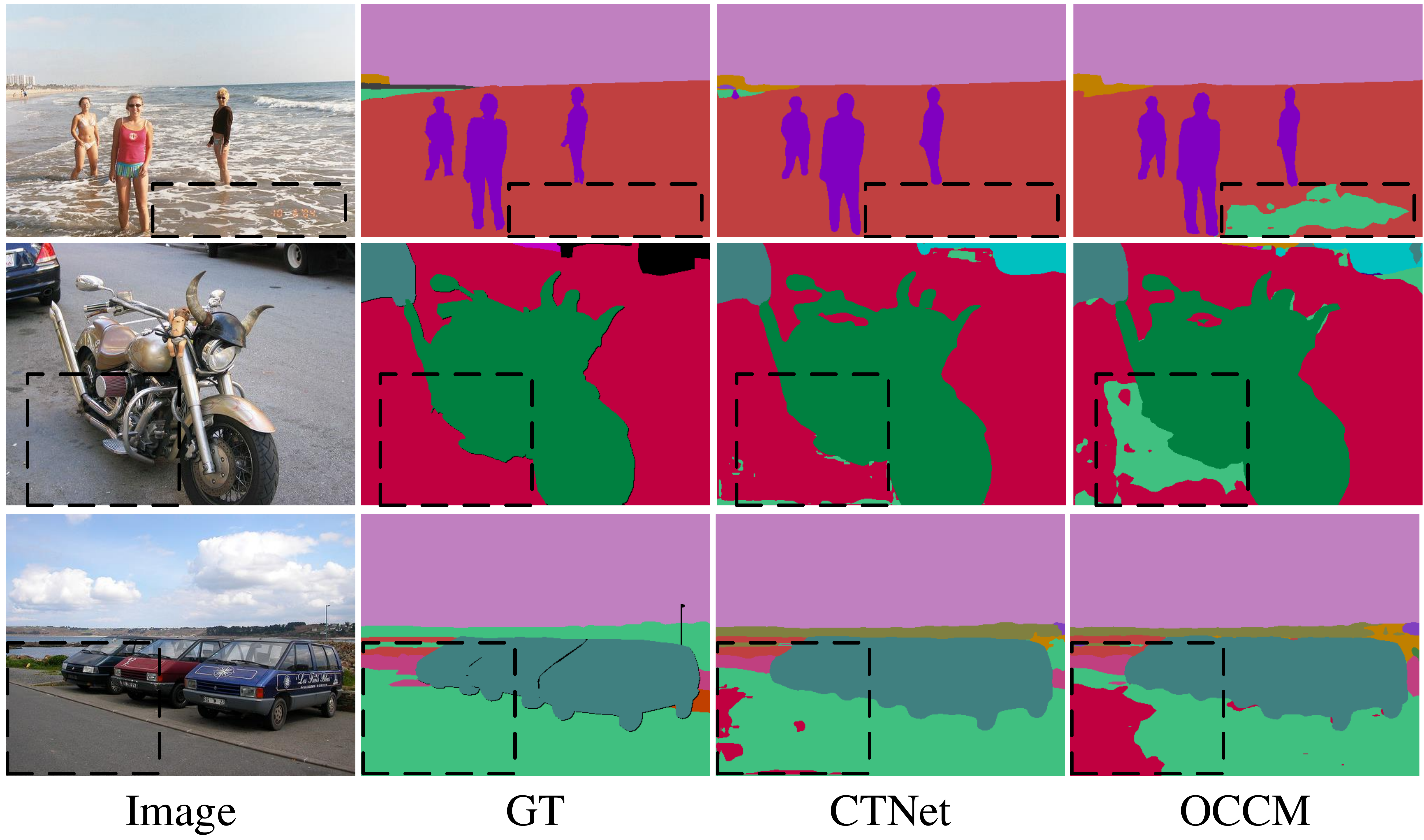}
	\caption{Illustration of segmented examples of CTNet and OCCM.}
	\label{fig8}
\end{figure}
	
\begin{figure}
\centering
	\includegraphics[width=1.0\linewidth]{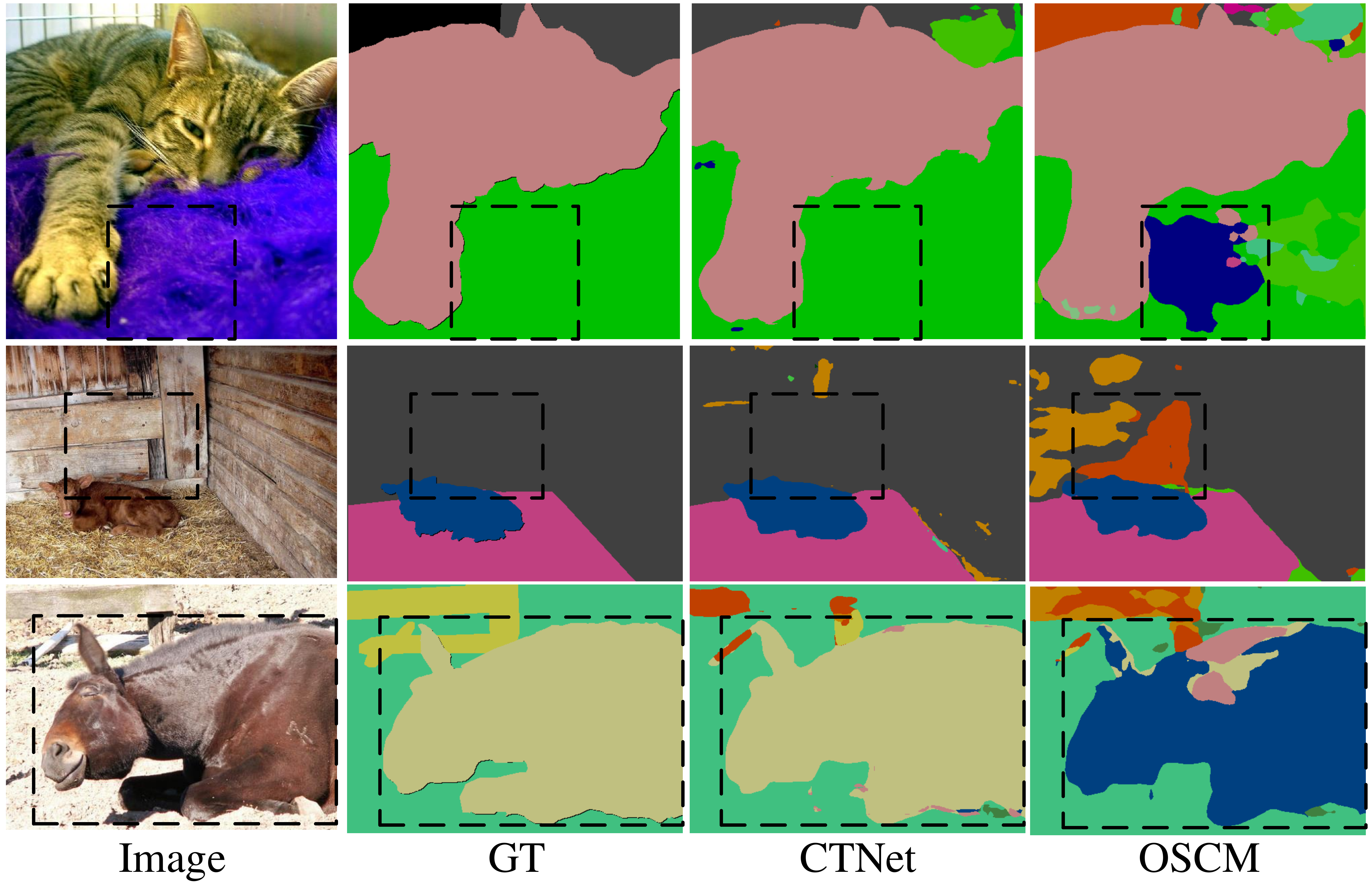}
	\caption{Illustration of segmented examples of CTNet and OSCM.}
	\label{fig9}
	\vspace{-4mm}
\end{figure}

\subsubsection{For single module and connection mode}
To better show the importance of the CCM and SCM modules in CTNet for semantic segmentation, the simplified versions of the proposed CTNet, termed as OCCM and OSCM are designed. OCCM only uses the CCM module as the segmentation head by removing the SCM module from CTNet. The quantitative comparison of OCCM, OSCM and CTNet is shown in Table \ref{table5}. Results based on ResNet-50 and ResNet-100 as the backbone in both single-scale and multi-scale tests are presented. Compared with OCCM, CTNet achieves better performance in all the cases, which indicate the necessity of exploring the spatial context. Similarly, it is reasonable to explore the channel context by comparing CTNet with OSCM. In a word, the motivation of jointly exploring the contexts of both dimensions can be well verified.
	
Besides, some illustrative segmentation results are shown in Figure \ref{fig8} and Figure \ref{fig9} to present the visualization segmentation results of OCCM, OSCM and CTNet. From the compared results between OCCM and CTNet in Figure \ref{fig8}, it can be observed that some pixels are incorrectly annotated by OCCM while the category labels can be accurately predicted. Because the channel context provides specific category information in the image, while incorrectly captures correlations between pixels. From Figure \ref{fig9}, we can see that OSCM can correctly aggregate pixels in the same object, while the phenomenon of category prediction errors occurs. This indicates that the spatial context can capture correlations between pixels and aggregate pixels from the same object, while it may predict the category labels incorrectly. Consequently, it is necessary to simultaneously capture the channel and spatial context to make them promote each other.
	
\begin{table}[t]
	\caption{Ablation study of connection mode in CTNet on PASCAL-Context.}
	\label{table6}
		\renewcommand\arraystretch{1.3}
		\centering
		\begin{tabular}{c|c|c|c}
			\hline
			Method &Backbone&SS mIoU(\%) &MS mIoU(\%)\\
			\hline
			DANet \cite{fu2019dual}&ResNet-50&-&50.1\\
			PANet & ResNet-50& 51.4&52.6\\
			CTNet& ResNet-50&51.7&\textbf{53.3}\\
			\hline
			DANet \cite{fu2019dual} &ResNet-101 &-&52.6\\
			PANet & ResNet-101 &53.8&55.4\\
			CTNet& ResNet-101&54.4&\textbf{56.2}\\
			\hline
	\end{tabular}
\end{table}

\begin{figure}
		\centering
		\includegraphics[width=\linewidth]{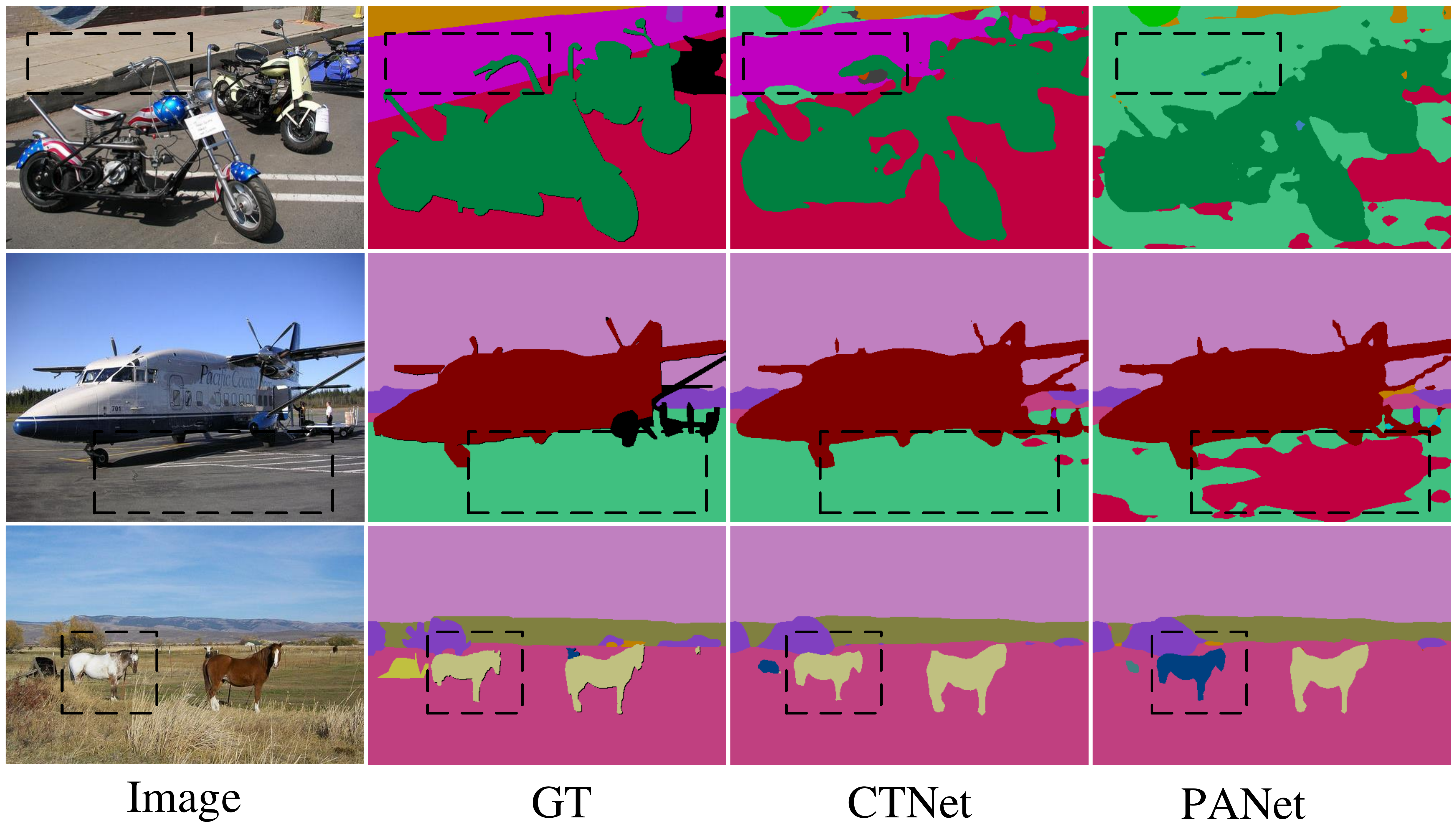}
		\caption{Illustration of segmented examples of CTNet and PANet.}
		\label{fig10}
\vspace{-4mm}
\end{figure}

Actually, there are different connection modes to simultaneously explore the channel and spatial context. Now experiments are conducted to evaluate the effectiveness of connection ways. There are two ways to connect two independent modules. That is, one is in tandem and the other is in parallel. Hence, the proposed tandem connection model is compared with the parallel connection model used in DANet. The quantitative results are reported in Table \ref{table6}. Besides, the revised version PANet of the proposed CTNet is compared, which explores the developed CCM and SCM in parallel. It can be seen that CTNet and PANet are better than DANet, which can show that the developed CCM and SCM enable to capture more accurate contexts. From the comparison between CTNet and PANet, we can see that it is reasonable to explore the channel and spatial contexts in tandem. Furthermore, some visualization results of CTNet and PANet are illustrated in Figure \ref{fig10}. Regions in the black box are incorrectly segmented by PANet. The results can indicate that the interactive learning of the channel context and the spatial context in tandem can improve the segmentation results.

\begin{figure}
		\centering
		\includegraphics[width=\linewidth]{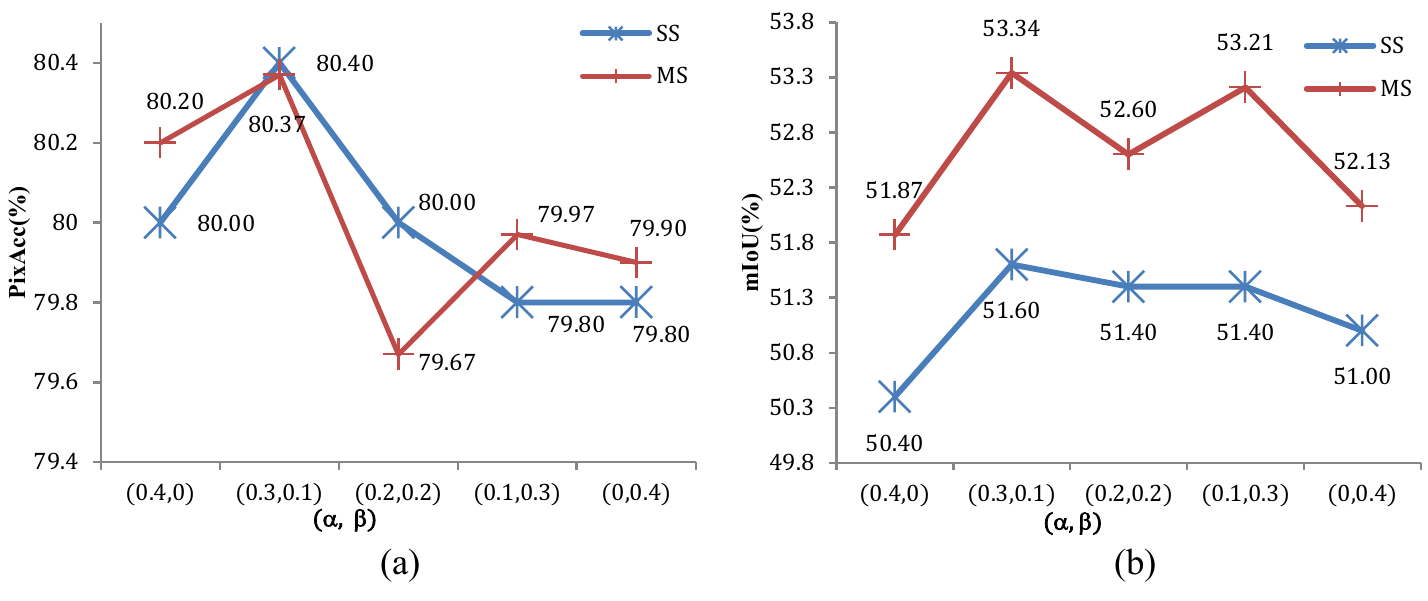}
		\caption{Ablation study of CP-Loss in CTNet on PASCAL-Context. $\alpha$ denotes the weight of auxiliary loss, and $\beta$ denotes the weight of CP-Loss.}
		\label{fig11}
\end{figure}

\subsubsection{For Class Probability Loss (CP-Loss)}
The proposed CTNet leverages the new loss function, i.e., the class probability loss, to improve the performance. The importance of CP-Loss is evaluated in this section. The results are shown in Figure \ref{fig11}. When $\beta=0$, that is, the CP-Loss is not used in the proposed CTNet, the performance in terms of mIoU is the worst, which can demonstrate that the channel context obtained by CCM has some noise information which affects the performance of CTnet. With the addition of CP-loss, CCM can selectively enhance the category-related feature maps and provide more accurate prior information for SCM. Besides, the best results are obtained when $\alpha=0.3$ and $\beta=0.1$. As the weight of cp-loss increases, CTNet performance tends to increase first and then decrease.Indicates that when CP-loss weights too much, CTnet places too much emphasis on channel context, thus breaking the balance between channel context and spatial context. Besides, when $\alpha=0.3$ and $\beta=0.1$, CTnet achieves optimal performance, indicating that the channel context and spatial context in the model are balanced.

\begin{table}[!t]
		\caption{Comparison in terms of mIoU on 60 classes from PASCAL-Context with state-of-the-arts.}
		\label{table7}
		\renewcommand\arraystretch{1.3}
		\centering
		\begin{tabular}{c|c|c}
			\hline
			Method &Backbone  & mIoU (\%)\\
			\hline
			FCN \cite{shelhamer2017fully}   & -  &   37.8\\
			VeryDeep \cite{wang2017learning} & - & 44.5\\
			RefineNet \cite{lin2017refinenet} & ResNet-152& 47.3\\
			PSPNet \cite{zhao2017pyramid} & ResNet-101& 47.8\\
			MSCI \cite{lin2018multi} & ResNet-152& 50.3\\
			EncNet \cite{zhang2018context}& ResNet-101& 51.7\\
			DANet \cite{fu2019dual}& ResNet-101& 52.6\\
			FastFCN \cite{wu2019fastfcn}& ResNet-101& 53.1\\
			EMANet \cite{li2019expectation}& ResNet-101& 53.1\\
			SVCNet \cite{ding2019semantic} & ResNet-101& 53.2\\
			CPN \cite{yu2020context}& ResNet-101&53.9\\
			CFNet \cite{zhang2019co}& ResNet-101& 54.0\\
			ACNet \cite{ding2019acnet} & ResNet-101& 54.1\\
			SPNet \cite{hou2020strip}& ResNet-101& 54.5\\
			APCNet \cite{he2019adaptive}& ResNet-101& 54.7\\
			OCR \cite{yuan2020object}& ResNet-101& 54.8\\
			\hline
			CTNet & ResNet-50&52.9\\
			CTNet & ResNet-101& \textbf{55.5}\\
			\hline
		\end{tabular}
\end{table}

\begin{figure}
		\centering
		\includegraphics[width=\linewidth]{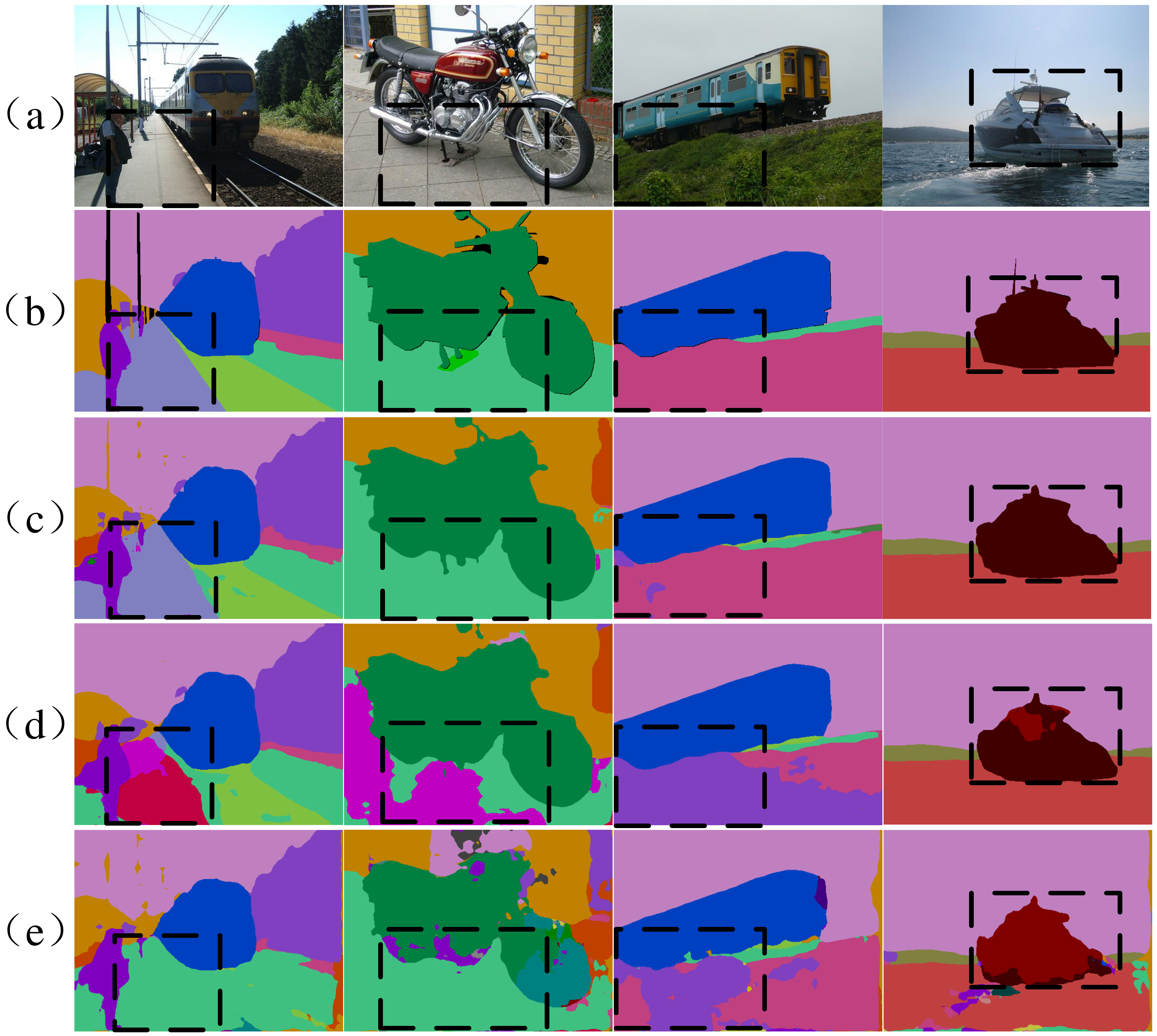}
		\caption{Visualization results of EncNet, FastFCN, and CTNet based on ResNet-101 on PASCAL-Context. (a) input image, (b) GT, (c) CTNet, (d) EncNet, and (e) FastFCN.}
		\label{fig12}
\vspace{-4mm}
\end{figure}

\subsection{State-of-the-Art Comparisons}
This section conducts experiments to compare the proposed CTNet with some state-of-the-arts. Results of the compared methods are from the original papers.
	
\subsubsection{Results on Pascal-Context}
Following previous work \cite{yuan2020object, ding2019acnet}, the proposed CTNet is compared with state-of-the-art methods based on $60$ categories in the Pascal-Context dataset. The compared methods includes FCN \cite{shelhamer2017fully}, VeryDeep \cite{wang2017learning}, RefineNet \cite{lin2017refinenet}, PSPNet \cite{zhao2017pyramid}, MSCI \cite{lin2018multi}, EncNet \cite{zhang2018context}, DANet \cite{fu2019dual}, FastFCN \cite{wu2019fastfcn}, SVCNet \cite{ding2019semantic}, CPN \cite{yu2020context}, CFNet \cite{zhang2019co}, ACNet \cite{ding2019acnet}, SPNet \cite{hou2020strip}, APCNet \cite{he2019adaptive} and OCR \cite{yuan2020object}. The compared results are presented on Table \ref{table7}. It can be easily observed that CTNet achieves the best result 55.5\% mIoU on PASCAL-Context test, which surpasses all the compared methods with a large margin even without using COCO pre-training. Second, the proposed CTNet even using ResNet-50 as backbone performs better than RefineNet \cite{lin2017refinenet} and MSCI \cite{lin2018multi} that employ a deeper network. Furthermore, when using the same backbone, CTNet performs significantly better than approaches that utilize a single context and DANet that uses the channel and spatial contexts. Because CTNet interactively both contexts for semantic segmentation.

To visualize the superiority of CTNet, qualitative comparisons between EncNet \cite{zhang2018context}, FastFCN \cite{wu2019fastfcn} and the proposed CTNet are presented in Figure \ref{fig12}. It can be observed that EncNet and FastFCN can't distinguish confused categories in image, while CTNet enables to distinguish them and generate better segmentation results. For example, EncNet and FastFCN incorrectly identifies some pixels in the third image belonging to\textit{'grass'} as \textit{'tree'} since the region of \textit{'grass'} is very similar in color and shape to the region of \textit{'tree'}. Fortunately, the proposed CTNet enables to correctly predict their label.
	
\begin{table}[t]
		\caption{Comparison with state-of-the-arts on the ADE20K val set.}
		\label{table8}
		\renewcommand\arraystretch{1.3}
		\centering
		\begin{tabular}{c|c|c}
			\hline
			Method &Backbone  & mIoU (\%)\\
			\hline
			RefineNet \cite{lin2017refinenet}& ResNet-152& 40.70\\
			UperNet \cite{xiao2018unified}& ResNet-101& 42.66\\
			PSPNet \cite{zhao2017pyramid}& ResNet-152& 43.51\\
			DSSPN \cite{liang2018dynamic}& ResNet-101& 43.68\\
			SAC \cite{zhang2017scale}& ResNet-101& 44.30\\
			FastFCN \cite{wu2019fastfcn} & ResNet-101& 44.34\\
			EncNet \cite{zhang2018context} & ResNet-101& 44.65\\
			GCU \cite{li2018beyond}& ResNet-101& 44.81\\
			CFNet \cite{zhang2019co}& ResNet-101& 44.89\\
			ALNN  \cite{zhu2019asymmetric}&ResNet-101&45.24\\
			OCR \cite{yuan2020object}&ResNet-101&45.28\\
			OCR \cite{yuan2020object}&HRNet-W48&45.66\\
			ACNet\cite{fu2019adaptive}&ResNet-101&45.90\\
			\hline
			CTNet & ResNet-101& \textbf{45.94}\\
			\hline
	\end{tabular}
\end{table}
	
\subsubsection{Results on ADE20K}
The ADE20K dataset is a challenging dataset for semantic segmentation. To show the effectiveness of the proposed CTNet, experiments are carried out compared with RefineNet \cite{lin2017refinenet}, UperNet \cite{xiao2018unified}, PSPNet \cite{zhao2017pyramid}, DSSPN \cite{liang2018dynamic}, SAC \cite{zhang2017scale}, FastFCN \cite{wu2019fastfcn}, EncNet \cite{zhang2018context}, GCU \cite{li2018beyond}, CFNet \cite{zhang2019co}, ALNN \cite{zhu2019asymmetric}, OCR \cite{yuan2020object} and ACNet\cite{fu2019adaptive}. Following previous work, the performance is evaluated on the validation set. From the compared results in Table \ref{table8}, CTNet achieves the best performance 45.94\% mIoU. The observations from the Pascal-Context dataset can be also obtained. It is worth noting that CTNet is better than OCR \cite{yuan2020object} using a more advanced backbone.

\begin{figure*}
	\centering
	\includegraphics[width=0.95\textwidth]{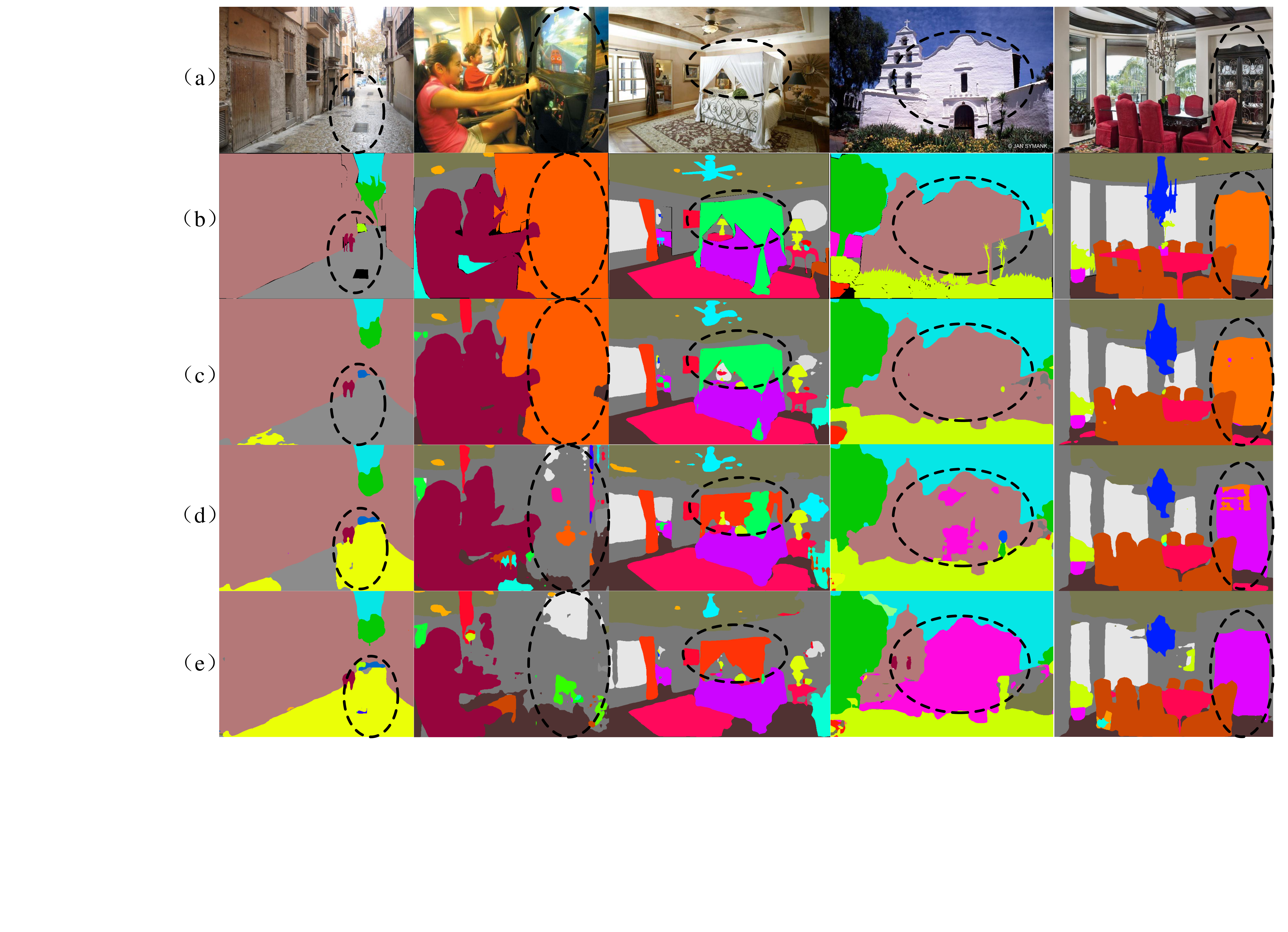}
		\caption{Visualization results of EncNet, FastFCN, and CTNet on ADE20K. (a) input image, (b) GT, (c) CTNet, (d) EncNet, and (e) FastFCN.}
	\label{fig13}
\vspace{-4mm}
\end{figure*}

Besides, some segmentation examples are illustrated in Figure \ref{fig13}. To show the advantages of CTNet more intuitively, we use black circles to mark the most challenging category information. We can see that EncNet and FastFCN easily mislabel pixels in the black circle, while CTNet can label these pixels accurately. For example, EncNet incorrectly labels the regions of \textit{'cabinet'} and \textit{'bed'} in the fifth image, while CTNet accurately distinguishes them. Because CTNet explores the semantic dependency of channel and spatial features to enhance the network representation ability, and obtain good segmentation results.

\subsubsection{Results on Pascal VOC2012}
Experiments are also conducted on Pascal VOC2012 for semantic segmentation to compare the proposed CTN with FCN \cite{shelhamer2017fully}, Deeplabv2 \cite{chen2017deeplab}, CRF-RNN \cite{zheng2015conditional}, DeconvNet \cite{noh2015learning}, GCRF \cite{vemulapalli2016gaussian}, DPN \cite{liu2015semantic}, Piecewise \cite{lin2016efficient}, AFF \cite{ke2018adaptive}, ResNet38 \cite{wu2019wider}, DANet \cite{fu2019dual}, PSPNet \cite{zhao2017pyramid}, DFN \cite{yu2018learning}, EncNet \cite{zhang2018context}, PAN \cite{li2018pyramid}, APCNet \cite{he2019adaptive}, CFNet \cite{zhang2019co}, OCR \cite{yuan2020object} and DMNet \cite{he2019dynamic}. CTNet is first trained on the enhanced dataset \cite{hariharan2015hypercolumns}, and then fine-turned on the original PASCAL VOC2012 dataset. The segmentation results are obtained by submitting the test results to the official evaluation server. The performance over each category and mIoU are presented in Table \ref{table9}. Without COCO pre training, the proposed CTNet gains the best result $85.3\%$ mIoU. The proposed tandem network can make full use of the context information on two dimensions since the spatial context and channel context modules communicate with each other. 

\begin{table*}[t]
		\scriptsize
		\centering
		\caption{The compared segmentation performance over each category and mIoU on PASCAL VOC2012. }
		\label{table9}
		\renewcommand\arraystretch{1.3}
		\begin{tabular}{c|c|c|c|c|c|c|c|c|c|c|c|c|c|c|c|c|c|c|c}
			\hline
			& \rotatebox{90}{FCN \cite{shelhamer2017fully}} & \rotatebox{90}{Deeplabv2 \cite{chen2017deeplab}} & \rotatebox{90}{CRF-RNN \cite{zheng2015conditional}} & \rotatebox{90}{DeconvNet \cite{noh2015learning}} & \rotatebox{90}{GCRF \cite{vemulapalli2016gaussian}} & \rotatebox{90}{DPN \cite{liu2015semantic}} & \rotatebox{90}{Piecewise \cite{lin2016efficient}} & \rotatebox{90}{AFF \cite{ke2018adaptive}} & \rotatebox{90}{ResNet38 \cite{wu2019wider}} & \rotatebox{90}{DANet \cite{fu2019dual}} & \rotatebox{90}{PSPNet \cite{zhao2017pyramid}} & \rotatebox{90}{DFN \cite{yu2018learning}} & \rotatebox{90}{EncNet \cite{zhang2018context}} & \rotatebox{90}{PAN \cite{li2018pyramid}} & \rotatebox{90}{APCNet \cite{he2019adaptive}} & \rotatebox{90}{CFNet \cite{zhang2019co}} & \rotatebox{90}{OCR \cite{yuan2020object}} & \rotatebox{90}{DMNet \cite{he2019dynamic}} & \rotatebox{90}{CTNet} \\
			\hline
			aero & 76.8 & 84.4 & 87.5 & 89.9 & 85.2 & 87.7 & 90.6 & 91.2 & 94.4 & - & 91.8 & - & 94.1 & 95.7 & 95.8 & 95.7 & - & \textbf{96.1} & \textbf{96.1} \\
			
			bike & 34.2 & 54.5 & 39.0 & 39.3 & 43.9 & 59.4 & 37.6 & 72.9 & 72.9 & - & 71.9 & - & 69.2 & 75.2 & 75.8 & 71.9 & - & \textbf{77.3} & 75.9 \\
			
			bird & 68.9 & 81.5 & 79.7 & 79.9 & 73.3 & 78.4 & 80.0 & 90.7 & 94.9 & - & 94.7 & - & 96.3 & 94.0 & 84.5 & 95.0 & - & 94.1 & \textbf{96.8} \\
			
			boat & 49.4 & 63.6 & 64.2 & 63.9 & 65.2 & 64.9 & 67.8 & 68.2 & 68.8 & - & 71.2 & - & 76.7 & 73.8 & 76.0 & 76.3 & - & 72.8 & \textbf{78.0} \\
			
			bottle & 60.3 & 65.9 & 68.3 & 68.2 & 68.3 & 70.3 & 74.4 & 77.7 & 78.4 & - & 75.8 & - & \textbf{86.2} & 79.6 & 80.6 & 82.8 & - & 78.1 & 82.4 \\
			
			bus & 75.3 & 85.1 & 87.6 & 87.4 & 89.0 & 89.3 & 92.0 & 95.6 & 90.6 & - & 95.2 & - & 96.3 & 96.5 & 96.9 & 94.8 & - & \textbf{97.1} & 95.3 \\
			
			car & 74.7 & 79.1 & 80.8 & 81.2 & 82.7 & 83.5 & 85.2 & 90.7 & 90.0 & - & 89.9 & - & 90.7 & \textbf{93.7} & 90.0 & 90.0 & - & 92.7 & 92.3 \\
			
			cat & 77.6 & 83.4 & 84.4 & 86.1 & 85.3 & 86.1 & 86.2 & 94.7 & 92.1 & - & 95.9 & - & 94.2 & 94.1 & 96.0 & 95.9 & - & 96.4 & \textbf{96.7} \\
			
			chair & 21.4 & 30.7 & 30.4 & 28.5 & 31.1 & 31.7 & 39.1 & 40.9 & 40.1 & - & 39.3 & - & 38.8 & 40.5 & \textbf{42.0} & 37.1 & - & 39.8 & \textbf{42.0} \\
			
			cow & 62.5 & 74.1 & 78.2 & 77.0 & 79.5 & 79.9 & 81.2 & 89.5 & 90.4 & - & 90.7 & - & 90.7 & 93.3 & 93.7 & 92.6 & - & 91.4 & \textbf{93.8} \\
			
			table & 46.8 & 59.8 & 60.4 & 62.0 & 63.3 & 62.6 & 58.9 & 72.6 & 71.7 & - & 71.7 & - & 73.3 & 72.4 & 75.4 & 73.0 & - & \textbf{75.5} & 71.2 \\
			
			dog & 71.8 & 79.0 & 80.5 & 79.0 & 80.5 & 81.9 & 83.8 & 91.6 & 89.9 & - & 90.5 & - & 90.0 & 89.1 & 91.6 & 93.4 & - & 92.7 & \textbf{93.8} \\
			
			horse & 63.9 & 76.1 & 77.8 & 80.3 & 79.3 & 80.0 & 83.9 & 94.1 & 93.7 & - & 94.5 & - & 92.5 & 94.1 & 95.0 & 94.6 & - & \textbf{95.8} & 95.0 \\
			
			mbike & 76.5 & 83.2 & 83.1 & 83.6 & 85.5 & 83.5 & 84.3 & 88.3 & 91.0 & - & 88.8 & - & 88.8 & \textbf{91.6} & 90.5 & 89.6 & - & 91.0 & 90.5 \\
			
			person & 73.9 & 80.8 & 80.6 & 80.2 & 81.0 & 82.3 & 84.8 & 88.8 & 89.1 & - & 89.6 & - & 87.9 & 89.5 & 89.3 & 88.1 & - & 90.3 & \textbf{90.6} \\
			
			plant & 45.2 & 59.7 & 59.5 & 58.8 & 60.5 & 60.5 & 62.1 & 67.3 & 71.3 & - & 72.8 & - & 68.7 & 73.6 & 75.8 & 74.9 & - & 76.6 & \textbf{77.9} \\
			
			sheep & 72.4 & 82.2 & 82.8 & 83.4 & 95.5 & 93.2 & 93.2 & 92.9 & 90.7 & - & 89.6 & - & 92.6 & 93.2 & 92.8 & \textbf{95.2} & - & 94.1 & \textbf{95.2} \\
			
			sofa & 37.4 & 50.4 & 47.8 & 54.3 & 52.0 & 53.4 & 58.2 & 62.6 & 61.3 & - & \textbf{64.0} & - & 59.0 & 62.8 & 61.9 & 63.2 & - & 62.1 & 62.9 \\
			
			train & 70.9 & 73.1 & 78.3 & 80.7 & 77.3 & 77.9 & 80.8 & 85.2 & 87.7 & - & 85.1 & - & 86.4 & 87.3 & 88.9 & \textbf{89.7} & - & 85.5 & 89.5 \\
			
			tv & 55.1 & 63.7 & 67.1 & 65.0 & 65.1 & 65.0 & 72.3 & 74.0 & 78.1 & - & 76.3 & - & 73.4 & 78.6 & \textbf{79.6} & 78.2 & - & 77.6 & 78.4 \\
			\hline
			mIoU & 62.2 & 71.6 & 72.0 & 72.5 & 73.2 & 74.1 & 75.3 & 82.2 & 82.5 & 82.6 & 82.6 & 82.7 & 82.9 & 84.0 & 84.2 & 84.2 & 84.3 & 84.4 & \textbf{85.3} \\
			\hline
	\end{tabular}
\end{table*}
	
\section{Conclusion}
This paper proposes a Context-based Tandem Network for Semantic Segmentation (CTNet), which makes full use of the context information on two important dimensions of channel and spatial in images. Specifically, the Channel Contextual Model (CCM) and the Spatial Contextual Model (SCM) are developed to explore channel and spatial context respectively. At the same time, we connect these two modules in tandem for interactive training to realize the context information mutual communication between the two dimensions. The features learned by the CCM can be regarded as a prior knowledge to guide the SCM learning, which can improve the features of the SCM learning. The SCM module introduces a novel self-attention mechanism to improve the efficiency of the model without affecting its performance. CTNet achieves superior performance, i.e., 55.5\% mIoU on PASCAL-Context, 45.94\% mIoU on ADE20K and 85.3\% mIoU on PASCAL VOC2012, to the state-of-the-arts.


\bibliographystyle{IEEEtran}
\bibliography{refer}

\begin{thebibliography}{10}
\providecommand{\url}[1]{#1}
\csname url@samestyle\endcsname
\providecommand{\newblock}{\relax}
\providecommand{\bibinfo}[2]{#2}
\providecommand{\BIBentrySTDinterwordspacing}{\spaceskip=0pt\relax}
\providecommand{\BIBentryALTinterwordstretchfactor}{4}
\providecommand{\BIBentryALTinterwordspacing}{\spaceskip=\fontdimen2\font plus
\BIBentryALTinterwordstretchfactor\fontdimen3\font minus
  \fontdimen4\font\relax}
\providecommand{\BIBforeignlanguage}[2]{{%
\expandafter\ifx\csname l@#1\endcsname\relax
\typeout{** WARNING: IEEEtran.bst: No hyphenation pattern has been}%
\typeout{** loaded for the language `#1'. Using the pattern for}%
\typeout{** the default language instead.}%
\else
\language=\csname l@#1\endcsname
\fi
#2}}
\providecommand{\BIBdecl}{\relax}
\BIBdecl

\bibitem{hung2017scene}
W.-C. Hung, Y.-H. Tsai, X.~Shen, Z.~Lin, K.~Sunkavalli, X.~Lu, and M.-H. Yang,
  ``Scene parsing with global context embedding,'' in \emph{Proceedings of the
  IEEE International Conference on Computer Vision}, 2017, pp. 2631--2639.

\bibitem{chen2017rethinking}
L.-C. Chen, G.~Papandreou, F.~Schroff, and H.~Adam, ``Rethinking atrous
  convolution for semantic image segmentation,'' \emph{arXiv
  preprint:1706.05587}, 2017.

\bibitem{ding2018context}
H.~Ding, X.~Jiang, B.~Shuai, A.~Qun~Liu, and G.~Wang, ``Context contrasted
  feature and gated multi-scale aggregation for scene segmentation,'' in
  \emph{Proceedings of IEEE Conference on Computer Vision and Pattern
  Recognition}, 2018, pp. 2393--2402.

\bibitem{zhao2017pyramid}
H.~Zhao, J.~Shi, X.~Qi, X.~Wang, and J.~Jia, ``Pyramid scene parsing network,''
  in \emph{Proceedings of IEEE Conference on Computer Vision and Pattern
  Recognition}, 2017, pp. 2881--2890.

\bibitem{shen2020ranet}
D.~Shen, Y.~Ji, P.~Li, Y.~Wang, and D.~Lin, ``Ranet: Region attention network
  for semantic segmentation,'' \emph{Proceedings of Advances in Neural
  Information Processing Systems}, 2020.

\bibitem{xiao2018unified}
T.~Xiao, Y.~Liu, B.~Zhou, Y.~Jiang, and J.~Sun, ``Unified perceptual parsing
  for scene understanding,'' in \emph{Proceedings of European Conference on
  Computer Vision}, 2018, pp. 418--434.

\bibitem{zhang2020feature}
D.~Zhang, H.~Zhang, J.~Tang, M.~Wang, X.~Hua, and Q.~Sun, ``Feature pyramid
  transformer,'' in \emph{Proceedings of European Conference on Computer
  Vision}, 2020, pp. 323--339.

\bibitem{seyedhosseini2015semantic}
M.~Seyedhosseini and T.~Tasdizen, ``Semantic image segmentation with contextual
  hierarchical models,'' \emph{IEEE transactions on pattern analysis and
  machine intelligence}, vol.~38, no.~5, pp. 951--964, 2015.

\bibitem{ding2020semantic}
H.~Ding, X.~Jiang, B.~Shuai, A.~Q. Liu, and G.~Wang, ``Semantic segmentation
  with context encoding and multi-path decoding,'' \emph{IEEE Transactions on
  Image Processing}, vol.~29, pp. 3520--3533, 2020.

\bibitem{chen2014semantic}
L.-C. Chen, G.~Papandreou, I.~Kokkinos, K.~Murphy, and A.~L. Yuille, ``Semantic
  image segmentation with deep convolutional nets and fully connected crfs,''
  \emph{arXiv preprint:1412.7062}, 2014.

\bibitem{wang2018non}
X.~Wang, R.~Girshick, A.~Gupta, and K.~He, ``Non-local neural networks,'' in
  \emph{Proceedings of IEEE Conference on Computer Vision and Pattern
  Recognition}, 2018, pp. 7794--7803.

\bibitem{huang2019ccnet}
Z.~Huang, X.~Wang, L.~Huang, C.~Huang, Y.~Wei, and W.~Liu, ``Ccnet: Criss-cross
  attention for semantic segmentation,'' in \emph{Proceedings of IEEE
  International Conference on Computer Vision}, 2019, pp. 603--612.

\bibitem{li2019expectation}
X.~Li, Z.~Zhong, J.~Wu, Y.~Yang, Z.~Lin, and H.~Liu, ``Expectation-maximization
  attention networks for semantic segmentation,'' in \emph{Proceedings of IEEE
  International Conference on Computer Vision}, 2019, pp. 9167--9176.

\bibitem{fu2020contextual}
J.~Fu, J.~Liu, Y.~Li, Y.~Bao, W.~Yan, Z.~Fang, and H.~Lu, ``Contextual
  deconvolution network for semantic segmentation,'' \emph{Pattern
  Recognition}, vol. 101, p. 107152, 2020.

\bibitem{ni2019raunet}
Z.-L. Ni, G.-B. Bian, X.-H. Zhou, Z.-G. Hou, X.-L. Xie, C.~Wang, Y.-J. Zhou,
  R.-Q. Li, and Z.~Li, ``Raunet: Residual attention u-net for semantic
  segmentation of cataract surgical instruments,'' in \emph{Proceedings of
  International Conference on Neural Information Processing}, 2019, pp.
  139--149.

\bibitem{woo2018cbam}
S.~Woo, J.~Park, J.-Y. Lee, and I.~So~Kweon, ``Cbam: Convolutional block
  attention module,'' in \emph{Proceedings of European Conference on Computer
  Vision}, 2018, pp. 3--19.

\bibitem{hu2018squeeze}
J.~Hu, L.~Shen, and G.~Sun, ``Squeeze-and-excitation networks,'' in
  \emph{Proceedings of IEEE Conference on Computer Vision and Pattern
  Recognition}, 2018, pp. 7132--7141.

\bibitem{li2019selective}
X.~Li, W.~Wang, X.~Hu, and J.~Yang, ``Selective kernel networks,'' in
  \emph{Proceedings of IEEE Conference on Computer Vision and Pattern
  Recognition}, 2019, pp. 510--519.

\bibitem{zhang2018context}
H.~Zhang, K.~Dana, J.~Shi, Z.~Zhang, X.~Wang, A.~Tyagi, and A.~Agrawal,
  ``Context encoding for semantic segmentation,'' in \emph{Proceedings of IEEE
  Conference on Computer Vision and Pattern Recognition}, 2018, pp. 7151--7160.

\bibitem{wang2020eca}
Q.~Wang, B.~Wu, P.~Zhu, P.~Li, W.~Zuo, and Q.~Hu, ``Eca-net: Efficient channel
  attention for deep convolutional neural networks,'' in \emph{Proceedings of
  IEEE Conference on Computer Vision and Pattern Recognition}, 2020, pp.
  11\,534--11\,542.

\bibitem{fu2019dual}
J.~Fu, J.~Liu, H.~Tian, Y.~Li, Y.~Bao, Z.~Fang, and H.~Lu, ``Dual attention
  network for scene segmentation,'' in \emph{Proceedings of IEEE Conference on
  Computer Vision and Pattern Recognition}, 2019, pp. 3146--3154.

\bibitem{lin2017exploring}
G.~Lin, C.~Shen, A.~Van Den~Hengel, and I.~Reid, ``Exploring context with deep
  structured models for semantic segmentation,'' \emph{IEEE transactions on
  pattern analysis and machine intelligence}, vol.~40, no.~6, pp. 1352--1366,
  2017.

\bibitem{liu2015parsenet}
W.~Liu, A.~Rabinovich, and A.~C. Berg, ``Parsenet: Looking wider to see
  better,'' \emph{arXiv preprint:1506.04579}, 2015.

\bibitem{zhang2020causal}
D.~Zhang, H.~Zhang, J.~Tang, X.-S. Hua, and Q.~Sun, ``Causal intervention for
  weakly-supervised semantic segmentation,'' in \emph{Proceedings of Advances
  in Neural Information Processing Systems}, 2020.

\bibitem{zhang2019co}
H.~Zhang, H.~Zhang, C.~Wang, and J.~Xie, ``Co-occurrent features in semantic
  segmentation,'' in \emph{Proceedings of IEEE Conference on Computer Vision
  and Pattern Recognition}, 2019, pp. 548--557.

\bibitem{huang2019interlaced}
L.~Huang, Y.~Yuan, J.~Guo, C.~Zhang, X.~Chen, and J.~Wang, ``Interlaced sparse
  self-attention for semantic segmentation,'' \emph{arXiv preprint
  arXiv:1907.12273}, 2019.

\bibitem{vaswani2017attention}
A.~Vaswani, N.~Shazeer, N.~Parmar, J.~Uszkoreit, L.~Jones, A.~N. Gomez,
  {\L}.~Kaiser, and I.~Polosukhin, ``Attention is all you need,'' in
  \emph{Proceedings pf Advances in Neural Information Processing Systems},
  2017, pp. 5998--6008.

\bibitem{chen20182}
Y.~Chen, Y.~Kalantidis, J.~Li, S.~Yan, and J.~Feng, ``A\^{} 2-nets: Double
  attention networks,'' in \emph{Proceedings pf Advances in Neural Information
  Processing Systems}, 2018, pp. 352--361.

\bibitem{zhu2019asymmetric}
Z.~Zhu, M.~Xu, S.~Bai, T.~Huang, and X.~Bai, ``Asymmetric non-local neural
  networks for semantic segmentation,'' in \emph{Proceedings of IEEE
  International Conference on Computer Vision}, 2019, pp. 593--602.

\bibitem{yin2020disentangled}
M.~Yin, Z.~Yao, Y.~Cao, X.~Li, Z.~Zhang, S.~Lin, and H.~Hu, ``Disentangled
  non-local neural networks,'' in \emph{Proceedings of European Conference on
  Computer Vision}, 2020, pp. 191--207.

\bibitem{cao2019gcnet}
Y.~Cao, J.~Xu, S.~Lin, F.~Wei, and H.~Hu, ``Gcnet: Non-local networks meet
  squeeze-excitation networks and beyond,'' in \emph{Proceedings of IEEE
  International Conference on Computer Vision Workshops}, 2019.

\bibitem{lin2018multi}
D.~Lin, Y.~Ji, D.~Lischinski, D.~Cohen-Or, and H.~Huang, ``Multi-scale context
  intertwining for semantic segmentation,'' in \emph{Proceedings of European
  Conference on Computer Vision}, 2018, pp. 603--619.

\bibitem{yuan2020object}
Y.~Yuan, X.~Chen, and J.~Wang, ``Object-contextual representations for semantic
  segmentation,'' in \emph{Proceedings of European Conference on Computer
  Vision}, 2020, pp. 173--190.

\bibitem{mottaghi2014role}
R.~Mottaghi, X.~Chen, X.~Liu, N.-G. Cho, S.-W. Lee, S.~Fidler, R.~Urtasun, and
  A.~Yuille, ``The role of context for object detection and semantic
  segmentation in the wild,'' in \emph{Proceedings of IEEE Conference on
  Computer Vision and Pattern Recognition}, 2014, pp. 891--898.

\bibitem{zhou2017scene}
B.~Zhou, H.~Zhao, X.~Puig, S.~Fidler, A.~Barriuso, and A.~Torralba, ``Scene
  parsing through ade20k dataset,'' in \emph{Proceedings of IEEE Conference on
  Computer Vision and Pattern Recognition}, 2017, pp. 633--641.

\bibitem{everingham2010pascal}
M.~Everingham, L.~Van~Gool, C.~K. Williams, J.~Winn, and A.~Zisserman, ``The
  pascal visual object classes (voc) challenge,'' \emph{International Journal
  of Computer Vision}, vol.~88, no.~2, pp. 303--338, 2010.

\bibitem{hariharan2015hypercolumns}
B.~Hariharan, P.~Arbel{\'a}ez, R.~Girshick, and J.~Malik, ``Hypercolumns for
  object segmentation and fine-grained localization,'' in \emph{Proceedings of
  IEEE Conference on Computer Vision and Pattern Recognition}, 2015, pp.
  447--456.

\bibitem{he2016deep}
K.~He, X.~Zhang, S.~Ren, and J.~Sun, ``Deep residual learning for image
  recognition,'' in \emph{Proceedings of IEEE Conference on Computer Vision and
  Pattern Recognition}, 2016, pp. 770--778.

\bibitem{wu2019fastfcn}
H.~Wu, J.~Zhang, K.~Huang, K.~Liang, and Y.~Yu, ``Fastfcn: Rethinking dilated
  convolution in the backbone for semantic segmentation,'' \emph{arXiv
  preprint:1903.11816}, 2019.

\bibitem{shelhamer2017fully}
E.~Shelhamer, J.~Long, and T.~Darrell, ``Fully convolutional networks for
  semantic segmentation,'' \emph{IEEE transactions on pattern analysis and
  machine intelligence}, vol.~39, no.~4, pp. 640--651, 2017.

\bibitem{chen2018encoder}
L.-C. Chen, Y.~Zhu, G.~Papandreou, F.~Schroff, and H.~Adam, ``Encoder-decoder
  with atrous separable convolution for semantic image segmentation,'' in
  \emph{Proceedings of European Conference on Computer Vision}, 2018, pp.
  801--818.

\bibitem{yu2020context}
C.~Yu, J.~Wang, C.~Gao, G.~Yu, C.~Shen, and N.~Sang, ``Context prior for scene
  segmentation,'' in \emph{Proceedings of IEEE Conference on Computer Vision
  and Pattern Recognition}, 2020, pp. 12\,416--12\,425.

\bibitem{wang2017learning}
G.~Wang, P.~Luo, L.~Lin, and X.~Wang, ``Learning object interactions and
  descriptions for semantic image segmentation,'' in \emph{Proceedings of IEEE
  Conference on Computer Vision and Pattern Recognition}, 2017, pp. 5859--5867.

\bibitem{lin2017refinenet}
G.~Lin, A.~Milan, C.~Shen, and I.~Reid, ``Refinenet: Multi-path refinement
  networks for high-resolution semantic segmentation,'' in \emph{Proceedings of
  IEEE Conference on Computer Vision and Pattern Recognition}, 2017, pp.
  1925--1934.

\bibitem{ding2019semantic}
H.~Ding, X.~Jiang, B.~Shuai, A.~Q. Liu, and G.~Wang, ``Semantic correlation
  promoted shape-variant context for segmentation,'' in \emph{Proceedings of
  IEEE Conference on Computer Vision and Pattern Recognition}, 2019, pp.
  8885--8894.

\bibitem{ding2019acnet}
X.~Ding, Y.~Guo, G.~Ding, and J.~Han, ``Acnet: Strengthening the kernel
  skeletons for powerful cnn via asymmetric convolution blocks,'' in
  \emph{Proceedings of IEEE International Conference on Computer Vision}, 2019,
  pp. 1911--1920.

\bibitem{hou2020strip}
Q.~Hou, L.~Zhang, M.-M. Cheng, and J.~Feng, ``Strip pooling: Rethinking spatial
  pooling for scene parsing,'' in \emph{Proceedings of IEEE Conference on
  Computer Vision and Pattern Recognition}, 2020, pp. 4003--4012.

\bibitem{he2019adaptive}
J.~He, Z.~Deng, L.~Zhou, Y.~Wang, and Y.~Qiao, ``Adaptive pyramid context
  network for semantic segmentation,'' in \emph{Proceedings of IEEE Conference
  on Computer Vision and Pattern Recognition}, 2019, pp. 7519--7528.

\bibitem{liang2018dynamic}
X.~Liang, H.~Zhou, and E.~Xing, ``Dynamic-structured semantic propagation
  network,'' in \emph{Proceedings of IEEE Conference on Computer Vision and
  Pattern Recognition}, 2018, pp. 752--761.

\bibitem{zhang2017scale}
R.~Zhang, S.~Tang, Y.~Zhang, J.~Li, and S.~Yan, ``Scale-adaptive convolutions
  for scene parsing,'' in \emph{Proceedings of IEEE International Conference on
  Computer Vision}, 2017, pp. 2031--2039.

\bibitem{li2018beyond}
Y.~Li and A.~Gupta, ``Beyond grids: Learning graph representations for visual
  recognition,'' in \emph{Proceedings of Advances in Neural Information
  Processing Systems}, 2018, pp. 9225--9235.

\bibitem{fu2019adaptive}
J.~Fu, J.~Liu, Y.~Wang, Y.~Li, Y.~Bao, J.~Tang, and H.~Lu, ``Adaptive context
  network for scene parsing,'' in \emph{Proceedings of IEEE International
  Conference on Computer Vision}, 2019, pp. 6748--6757.

\bibitem{chen2017deeplab}
L.-C. Chen, G.~Papandreou, I.~Kokkinos, K.~Murphy, and A.~L. Yuille, ``Deeplab:
  Semantic image segmentation with deep convolutional nets, atrous convolution,
  and fully connected crfs,'' \emph{IEEE Transactions on Pattern Analysis and
  Machine Intelligence}, vol.~40, no.~4, pp. 834--848, 2017.

\bibitem{zheng2015conditional}
S.~Zheng, S.~Jayasumana, B.~Romera-Paredes, V.~Vineet, Z.~Su, D.~Du, C.~Huang,
  and P.~H. Torr, ``Conditional random fields as recurrent neural networks,''
  in \emph{Proceedings of IEEE International Conference on Computer Vision},
  2015, pp. 1529--1537.

\bibitem{noh2015learning}
H.~Noh, S.~Hong, and B.~Han, ``Learning deconvolution network for semantic
  segmentation,'' in \emph{Proceedings of IEEE International Conference on
  Computer Vision}, 2015, pp. 1520--1528.

\bibitem{vemulapalli2016gaussian}
R.~Vemulapalli, O.~Tuzel, M.-Y. Liu, and R.~Chellapa, ``Gaussian conditional
  random field network for semantic segmentation,'' in \emph{Proceedings of
  IEEE Conference on Computer Vision and Pattern Recognition}, 2016, pp.
  3224--3233.

\bibitem{liu2015semantic}
Z.~Liu, X.~Li, P.~Luo, C.-C. Loy, and X.~Tang, ``Semantic image segmentation
  via deep parsing network,'' in \emph{Proceedings of IEEE International
  Conference on Computer Vision}, 2015, pp. 1377--1385.

\bibitem{lin2016efficient}
G.~Lin, C.~Shen, A.~Van Den~Hengel, and I.~Reid, ``Efficient piecewise training
  of deep structured models for semantic segmentation,'' in \emph{Proceedings
  of IEEE Conference on Computer Vision and Pattern Recognition}, 2016, pp.
  3194--3203.

\bibitem{ke2018adaptive}
T.-W. Ke, J.-J. Hwang, Z.~Liu, and S.~X. Yu, ``Adaptive affinity fields for
  semantic segmentation,'' in \emph{Proceedings of European Conference on
  Computer Vision}, 2018, pp. 587--602.

\bibitem{wu2019wider}
Z.~Wu, C.~Shen, and A.~Van Den~Hengel, ``Wider or deeper: Revisiting the resnet
  model for visual recognition,'' \emph{Pattern Recognition}, vol.~90, pp.
  119--133, 2019.

\bibitem{yu2018learning}
C.~Yu, J.~Wang, C.~Peng, C.~Gao, G.~Yu, and N.~Sang, ``Learning a
  discriminative feature network for semantic segmentation,'' in
  \emph{Proceedings of IEEE Conference on Computer Vision and Pattern
  Recognition}, 2018, pp. 1857--1866.

\bibitem{li2018pyramid}
H.~Li, P.~Xiong, J.~An, and L.~Wang, ``Pyramid attention network for semantic
  segmentation,'' \emph{arXiv preprint arXiv:1805.10180}, 2018.

\bibitem{he2019dynamic}
J.~He, Z.~Deng, and Y.~Qiao, ``Dynamic multi-scale filters for semantic
  segmentation,'' in \emph{Proceedings of IEEE International Conference on
  Computer Vision}, 2019, pp. 3562--3572.

\end{thebibliography}

\begin{IEEEbiography}[{\includegraphics[width=1in,height=1.25in,clip,keepaspectratio]{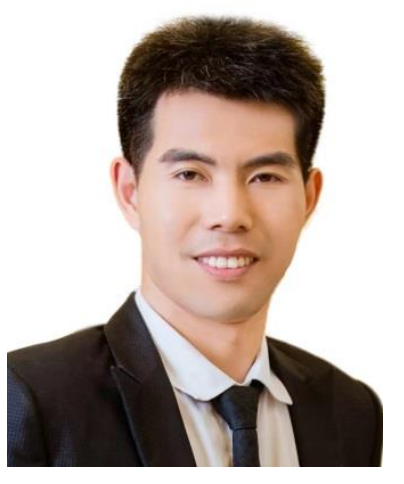}}] {Zechao Li} is currently a Professor at the Nanjing University of Science and Technology. He received his Ph.D degree from National Laboratory of Pattern Recognition, Institute of Automation, Chinese Academy of Sciences in 2013, and his B.E. degree from the University of Science and Technology of China in 2008. His research interests include big media analysis, computer vision, etc.
\end{IEEEbiography}
	
\begin{IEEEbiography}[{\includegraphics[width=1in,height=1.25in,clip,keepaspectratio]{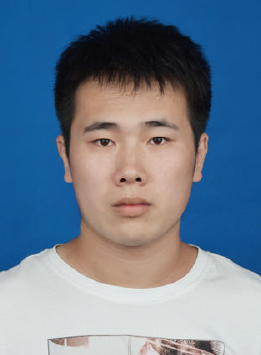}}] {Yanpeng Sun}received the MS degree at Guilin University Of Electronic Technology, China, in 2019. He is currently pursuing the Ph.D. degree with the School of Computer Science and Engineering, Nanjing University of Science and Technology, China. His research interests include deep learning, visual segmentation and understanding, etc.
\end{IEEEbiography}
	
\begin{IEEEbiography}[{\includegraphics[width=1in,height=1.25in,clip,keepaspectratio]{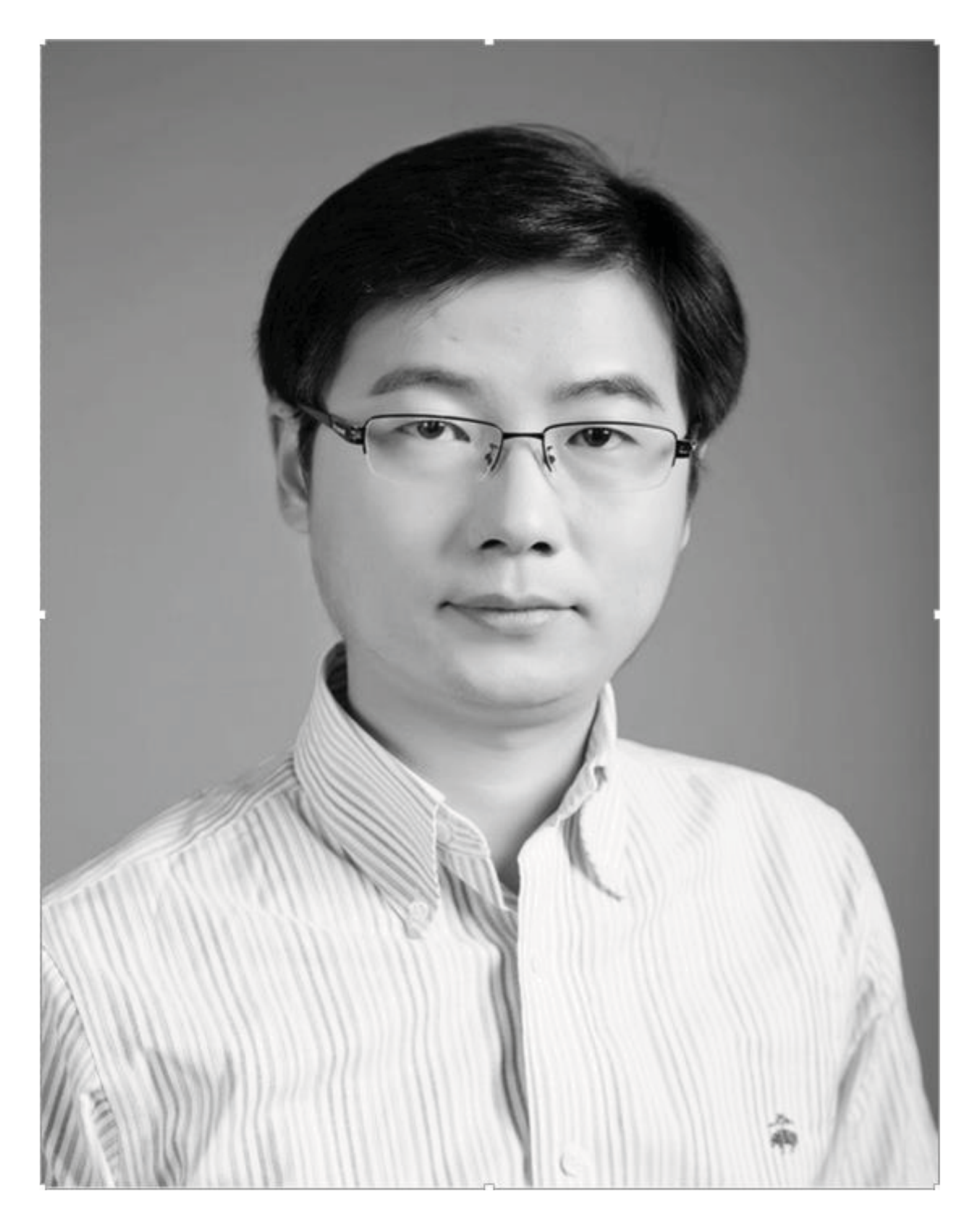}}] {Jinhui Tang} (M`08-SM`14) received the B.Eng. and Ph.D. degrees from the University of Science and Technology of China in 2003 and 2008, respectively. He is currently a Professor at the Nanjing University of Science and Technology. He has authored over 150 papers in top-tier journals and conferences. His research interests include multimedia analysis and computer vision. He was a recipient of the best paper awards in ACM MM 2007, PCM 2011 and ICIMCS 2011, the Best Paper Runner-up in ACM MM 2015, and the best student paper awards in MMM 2016 and ICIMCS 2017. He has served as an Associate Editor for the IEEE TNNLS, IEEE TKDE and IEEE TMM.
\end{IEEEbiography}

\end{document}